\documentclass[10pt,twocolumn,letterpaper]{article}
\usepackage{iccv}
\usepackage{amsmath,amssymb,amsfonts}
\usepackage{booktabs}
\usepackage{tabularx}
\usepackage{array}
\usepackage{float}
\usepackage{graphicx}
\usepackage{multirow}
\usepackage{makecell}
\usepackage[table]{xcolor}
\usepackage{caption}
\usepackage{subcaption}
\usepackage{cuted}
\usepackage[accsupp]{axessibility}
\newcolumntype{C}{>{\centering\arraybackslash}X}
\newcolumntype{L}{>{\raggedright\arraybackslash}X}
\newcolumntype{R}{>{\raggedleft\arraybackslash}X}

\newcommand{\TODO}[1]{}

\colorlet{blue}{black}

\definecolor{iccvblue}{rgb}{0.21,0.49,0.74}
\usepackage[pagebackref,breaklinks,colorlinks,allcolors=iccvblue]{hyperref}
\def\paperID{}
\def\confName{ICCV-CV4E}
\def\confYear{2025}
\title{TreeDGS: Aerial Gaussian Splatting for Distant DBH Measurement}
\author{
Belal Shaheen$^{1}$, Minh-Hieu Nguyen$^{1}$, Bach-Thuan Bui$^{1}$, Shubham$^{1}$, Tim Wu$^{1}$, Michael Fairley$^{1}$,\\
Matthew Zane$^{1}$, Michael Wu$^{1}$, James Tompkin$^{2}$\\
$^{1}$Coolant, San Francisco, CA 94111, USA\\
$^{2}$Department of Computer Science, Brown University, Providence, RI 02912, USA\\
{\tt\small michael@coolant.earth}
}
\begin{document}
\maketitle
\begin{strip}
\centering
\includegraphics[width=\textwidth]{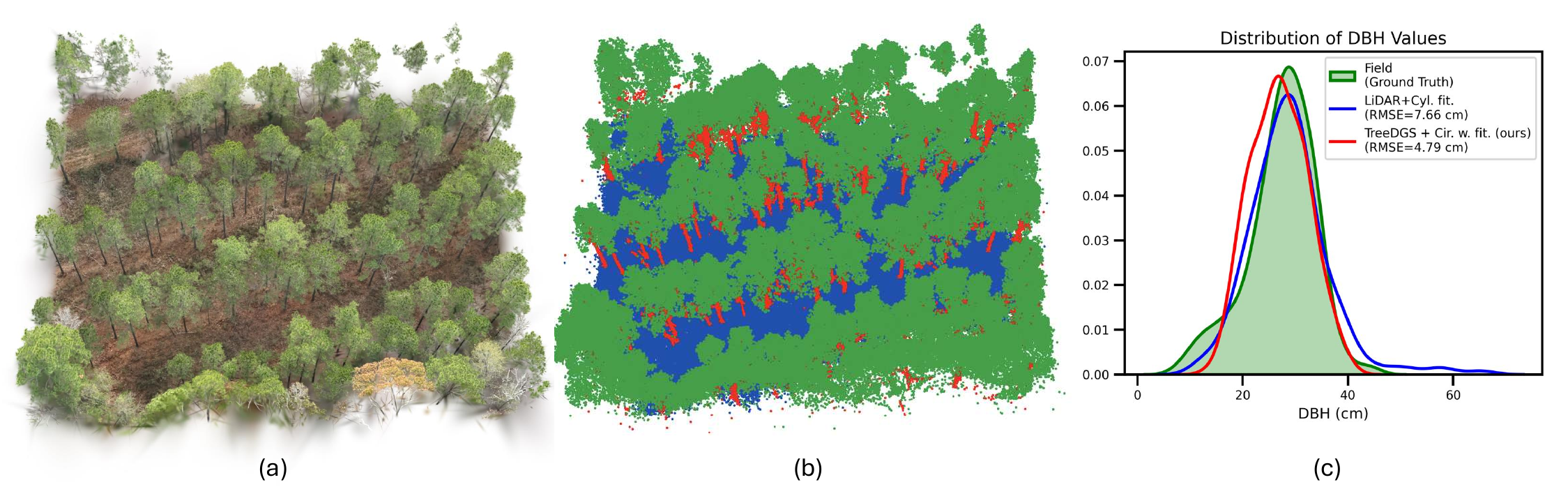}
\captionof{figure}{\textbf{TreeDGS accurately estimates DBH from distant RGB imagery:} (\textbf{a}) High-fidelity TreeDGS reconstruction from distant UAV RGB imagery as an optimized set of 3D Gaussians. (\textbf{b}) Surface points extracted via opacity-consistent sampling (built on RaDe-GS~\cite{zhang2024rade}) and segmented into stem vs. vegetation to isolate trunk geometry for DBH fitting. In (b), red points denote trunk/stem, green points denote vegetation/canopy, and blue points denote ground/other non-trunk points. (\textbf{c}) DBH errors against field measurements, shown as the distribution of DBH estimates relative to the field DBH distribution; TreeDGS + opacity-weighted circle fitting reduces error vs. UAV LiDAR + cylinder fitting~\cite{malladi2025digiforests} (RMSE/MAE: 4.79/3.67\,cm vs. 7.66/5.23\,cm) at a ground sample distance (GSD) of approximately 1.84~cm.}
\label{fig1}
\end{strip}
\begin{abstract}
Aerial remote sensing efficiently surveys large areas, but accurate direct object-level measurement remains difficult in complex natural scenes.
Advancements in 3D computer vision, particularly radiance field representations such as NeRF and 3D Gaussian splatting, can improve reconstruction fidelity from posed imagery.
Nevertheless, direct aerial measurement of important attributes like tree diameter at breast height (DBH) remains challenging. Trunks in aerial forest scans are distant and sparsely observed in image views; at typical operating altitudes, stems may span only a few pixels. With these constraints, conventional reconstruction methods have inaccurate breast-height trunk geometry.
TreeDGS is an aerial image reconstruction method that uses 3D Gaussian splatting as a continuous scene representation for trunk measurement.
After SfM--MVS initialization and Gaussian optimization, we extract a dense point set from the Gaussian field using RaDe-GS's depth-aware cumulative-opacity integration and associate each sample with a multi-view opacity reliability score. Then, we isolate trunk points and estimate DBH using opacity-weighted solid-circle fitting.
Evaluated on 10 plots with field-measured DBH, TreeDGS reaches {4.79 cm} 
 RMSE (about 2.6 pixels at this GSD) and outperforms a LiDAR baseline ({7.66 cm} RMSE). {This shows that TreeDGS can enable accurate, low-cost aerial DBH measurement (Figure \ref{fig1}).
}
\end{abstract}

\section{Introduction}
\label{sec:intro}

Aerial remote sensing has quickly become a cornerstone of modern environmental monitoring for its ability to survey large areas rapidly and cost {efficiently} 
~\cite{AndersonGaston2013UAVSpatialEcology,ColominaMolina2014UASReview,NexRemondino2014UAV3DReview}. Yet, despite major gains in sensor resolution, platform stability, and~reconstruction pipelines, extracting direct, object-level 
 measurements from aerial imagery remains difficult in structurally complex natural scenes~\cite{DandoisEllis2013RSE,Iglhaut2019SfMForestryReview}. Forested environments are particularly challenging as heterogeneous geometry, self-similar textures, frequent occlusion, and~strong appearance variability across lighting conditions all weaken the visual cues that aerial reconstruction methods rely on~\cite{Iglhaut2019SfMForestryReview}.

A prominent example is tree diameter at breast height (DBH). This describes an individual tree's size and growth stage. DBH is an important forestry variable that is a primary input to standard allometric models and reporting workflows to estimate wood volume, biomass, and~carbon~\cite{Brown1997,IPCC2006,Chave2014,Jenkins2003}. In turn, these allometric models underpin decision making used across forestry operations and policy, ranging from timber inventory and silvicultural planning to carbon offset quantification, wildfire risk mitigation, and~long-term ecosystem management~\cite{ScottReinhardt_CrownFire_2001, CARB_USForestProtocol_2015}.

\begin{figure}
    \includegraphics[width=0.99\linewidth]{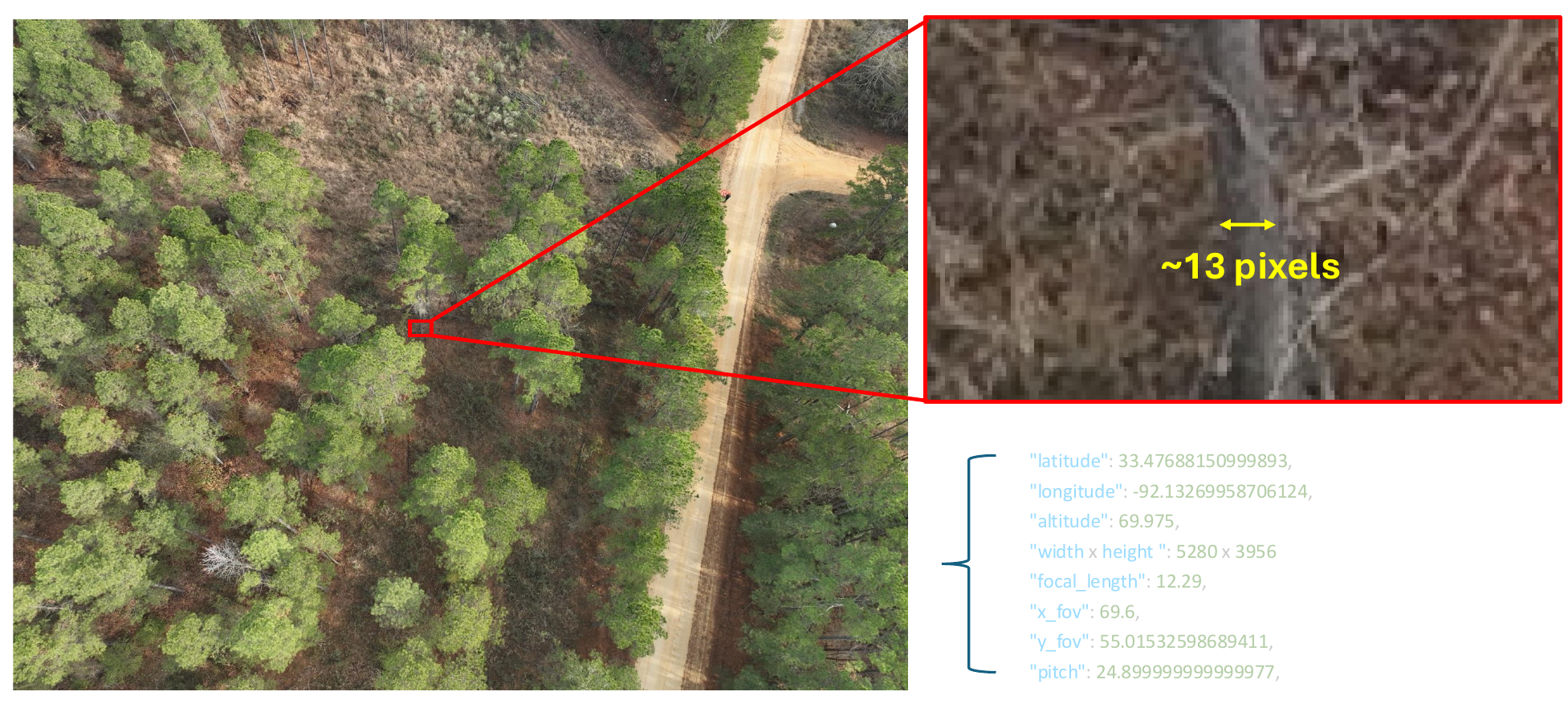}
    \caption{\textbf{{Pixel-limited trunk observations from distant aerial imagery.}} At \(\sim\)70\,m altitude \mbox{(\(\mathrm{GSD}\approx 1.84\)\,cm/px)}, a~typical pine stem can occupy only \(\sim\)13 pixels across in a single RGB image, making per-image diameter cues highly quantized, sensitive to occlusion, and~difficult to {measure~precisely.}}
    \label{fig:gsd_trunk}
\end{figure}

DBH estimation from above-canopy aerial sensing is fundamentally limited by observability. In~typical UAV image surveys flown at operational altitudes, stems are not large objects. In our data, captured at \(\sim\)70\,m above ground \mbox{(\(\mathrm{GSD}\approx 1.84\)\,cm/px)}, stems in the evaluation plots span only \(\sim\)15 pixels across per view on average (Table~\ref{tab:plot_summary}), and~they can be just a few pixels wide for smaller trees (Figure~\ref{fig:gsd_trunk}). This pixel scarcity is compounded by view scarcity: the breast-height band (\(h_{\mathrm{BH}}\approx 1.37\)\,m in our field protocol) is frequently occluded by crowns, branches, and~understory vegetation, so each stem may be seen only in a handful of oblique views through noisy canopy gaps. Thus, the~core geometric challenge of measurement is to recover enough faithful breast-height trunk surface samples to support stable circle or cylinder fitting.

Active sensing can mitigate some of these limitations by providing explicit range measurements rather than relying on image texture.~Terrestrial laser scanning (TLS), for~example, is well known to support robust DBH measurements because it can produce dense, well-populated stem cross-sections~\cite{Liang2016,Raumonen2013,Kankare2015}. However, the~mostly downward-looking trajectories of airborne or UAV laser scanning (ULS) do not typically yield equally dense returns on the trunk at breast height: even with high point density, trunk hits near \(h_{\mathrm{BH}}\) can be sparse, uneven, and~contaminated by surrounding vegetation, complicating per-tree stem isolation and reliable fitting in cluttered stands~\cite{Kuzelka2020,Neuville2021,Kukkonen2022}. These limitations help explain why UAV LiDAR has seen broader operational adoption for canopy structure and terrain products than for direct DBH measurement~\cite{Hollaus2007ALSStemVolume,Lim2003LidarForestStructure,Salas2010TreeDiameter,Wulder2008RoleLiDAR,Chen2017DTM}.

Nevertheless, RGB-only UAV photogrammetry has made substantial progress for other forest attributes, including canopy height models and structural mapping~\cite{DandoisEllis2013RSE,Iglhaut2019SfMForestryReview,White2013,Wallace2016}. In~contrast, DBH remains difficult from aerial imagery alone, even under careful survey design with aggressive overlap and strong oblique views~\cite{Moreira2021}. In~the pixel- and view-limited regime described above, conventional SfM--MVS pipelines often {yield incomplete and fragmented breast-height trunk surfaces that are insufficient for stable diameter fitting}. Sparse SfM relies on repeatable keypoints and long feature tracks~\cite{Schonberger2016}, but~distant trunks provide weak texture and are frequently interrupted by foliage occlusions, yielding unstable or missing correspondences. View-dependent MVS densification~\cite{Furukawa2010} then tends to drop out on trunks when visibility is intermittent and texture cues are weak. {Still, the~partial geometry recovered by MVS can serve as a useful coarse prior for later optimization}.

Recent radiance field representations, such as Neural Radiance Fields (NeRF)~\cite{Mildenhall2020} and 3D Gaussian Splatting (3DGS)~\cite{kerbl20233d}, can improve reconstruction fidelity from posed imagery and offer a promising path forward for thin structures that are only weakly observed in individual frames. Early forestry applications have begun exploring radiance fields for close-range terrestrial forest monitoring and tree-scale reconstruction~\cite{Huang2024,Korycki2025,shaheen2025forestsplat,Li2025Leafless}, but~the problem of direct DBH measurement from stand-off, above-canopy UAV surveys remains unsolved. Moreover, translating a ``fluffy'' radiance field reconstruction into a reliable diameter measurement requires an extraction step, such that we can densely sample trees at breast height and prioritize samples that are consistently supported across views.

We introduce a method---TreeDGS---that directly measures DBH from above-canopy UAV RGB imagery from a 3D Gaussian splatting scene reconstruction under pixel- and view-limited trunk observations. Starting from standard SfM camera poses and an MVS initialization, we optimize a RaDe-GS model~\cite{zhang2024rade} to obtain a continuous Gaussian field whose covariances and opacities are refined by multi-view photometric consistency. Then, we extract a dense point set using depth-aware cumulative-opacity integration, retaining samples whose accumulated opacity indicates a reliably occupied surface under a fixed threshold. To~mitigate spurious samples from foliage and partial occlusions, we compute a per-point multi-view opacity support score as a reliability signal and retain trunk-isolated points via 3D semantic segmentation. Finally, we estimate DBH with opacity-weighted, slice-wise circle fitting, yielding stable diameter estimates even when the trunk is only weakly observed in individual~images. We demonstrate real-world, field-validated performance in a managed loblolly pine forest with dense understory and varied growth patterns (10 plots), achieving 4.79\,cm RMSE (\(\sim\)2.6 pixels at our GSD) against ground-truth tape DBH and outperforming an ultra-high-resolution UAV LiDAR baseline by {37.5}\% in RMSE.

\section{Materials}
\label{sec:materials}

\begin{table}[t]
\centering
\scriptsize
\setlength{\tabcolsep}{4pt}
\renewcommand{\arraystretch}{0.98}
\caption{Per-plot summary for the manually paired subset used in per-tree evaluation. \#Trees denotes the number of stems in each plot with a verified field-to-reconstruction association; the ``All'' row reports pooled statistics across all plots.}
\label{tab:plot_summary}
\begin{tabular}{c c c c}
\toprule
\textbf{Plot} & \textbf{\#Trees} & \makecell[c]{\textbf{DBH Mean $\pm$ SD}\\\textbf{(cm)}} & \makecell[c]{\textbf{DBH Range}\\\textbf{(cm)}} \\
\midrule
1  & 22  & 31.67 $\pm$ 5.11 & 25.6--43.3 \\
2  & 27  & 28.23 $\pm$ 5.18 & 15.0--34.8 \\
3  & 25  & 26.26 $\pm$ 7.08 & 7.9--44.3  \\
4  & 16  & 24.52 $\pm$ 7.30 & 9.8--34.0  \\
5  & 22  & 28.18 $\pm$ 4.13 & 17.8--34.3 \\
6  & 23  & 27.02 $\pm$ 6.49 & 11.3--36.7 \\
7  & 19  & 20.87 $\pm$ 7.60 & 10.2--38.4 \\
8  & 17  & 29.25 $\pm$ 6.38 & 13.4--39.0 \\
9  & 19  & 27.32 $\pm$ 8.61 & 11.8--55.7 \\
10 & 20  & 29.53 $\pm$ 4.20 & 21.1--36.9 \\
All & 210 & 27.39 $\pm$ 6.73 & 7.9--55.7 \\
\bottomrule
\end{tabular}
\end{table}

\begin{figure}
    \includegraphics[width=\linewidth]{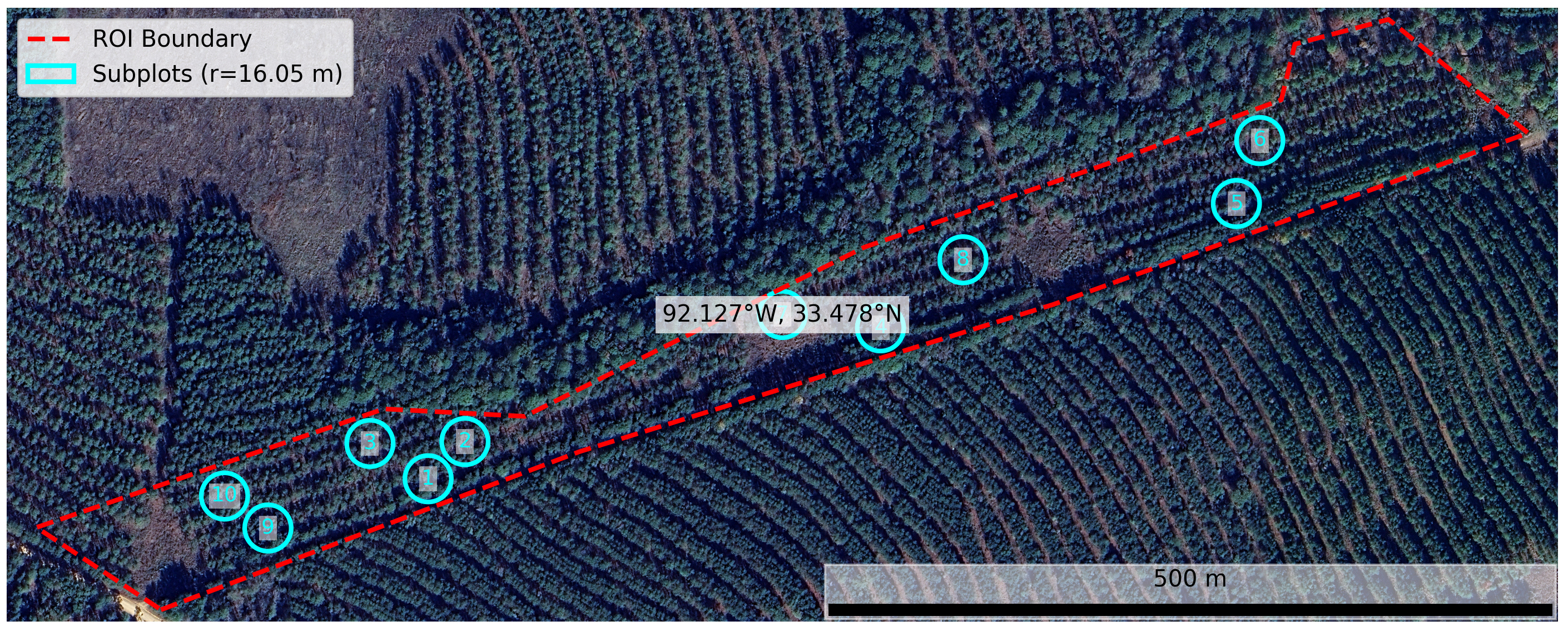}
    \caption{\textbf{Region of interest (ROI) and plot layout (10 subplots).} Each subplot corresponds to one 0.2-acre circular plot (radius 16.05\,m). The~circle indicates the field plot boundary used for tree inclusion, while the dashed polygon outlines the ROI used to clip and organize aerial products for per-plot~processing.}
    \label{fig:ROI}
\end{figure}

We collected data in a managed Pinus taeda (loblolly pine) stand in southeastern Arkansas, USA. We established ten 0.2-acre circular plots (plots~1--10; radius 16.05\,m; Figure~\ref{fig:ROI}). At each plot center, we marked a circular plot boundary with a 16.05\,m radius and measured all stems within the boundary.
Then, we assigned a unique ID to each included tree.
For each tree, we recorded the following:
(i) Bearing and distance from the plot center;
(ii) DBH measured with a diameter tape at 1.37\,m above ground level; and
(iii) height measured using a laser~rangefinder.

We surveyed plot centers using ORS/NTRIP network corrections. For~each plot, we centered a~survey-grade GNSS rover mounted on a survey pole and stabilized with a bipod over the plot and occupied for 30\,min in static mode while receiving RTCM corrections via an NTRIP caster. Then, we derived plot-center coordinates using a differential GNSS workflow, following forest-canopy positioning best-practice recommendations described by Strunk~et~al.~\cite{Strunk2025GNSS}. Finally, we exported corrected plot-center coordinates for co-registration with the aerial~products.

\begin{figure}[H]

    \includegraphics[width=\linewidth]{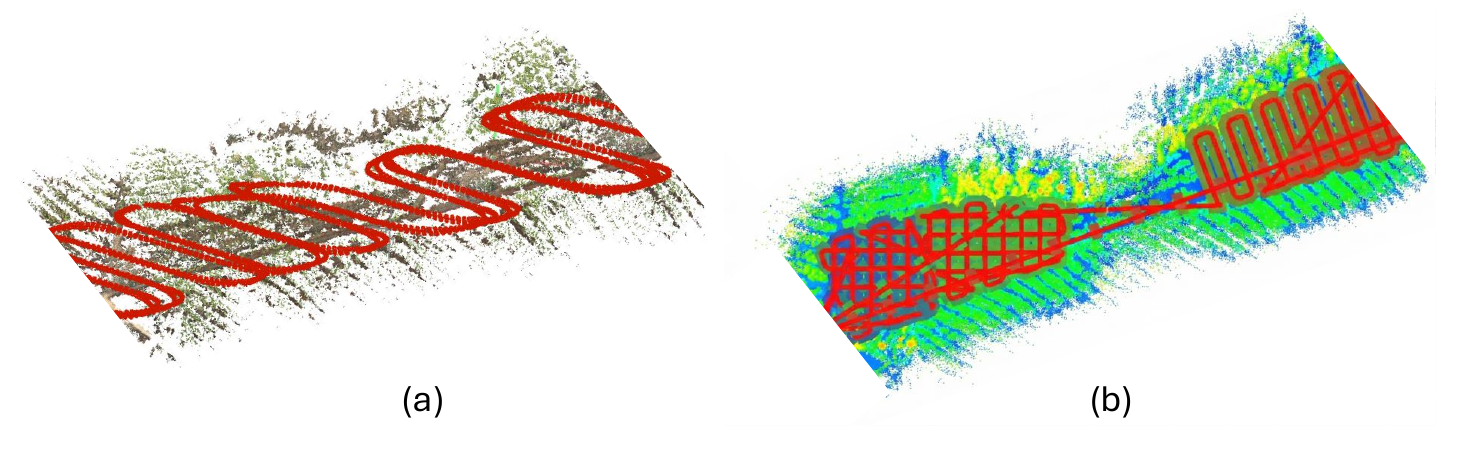}
    \caption{\textbf{{RGB and LiDAR acquisition patterns.}} (\textbf{a}) We collected RGB imagery with high overlap and mixed viewing angles to support SfM/MVS and improve trunk visibility. (\textbf{b}) We captured LiDAR with dense grid (lawnmower) flight lines to obtain uniform coverage across the {plot~network.} Red polylines indicate UAV flight trajectories; in (a) points are colored by RGB, while in (b) LiDAR points are colored by height above ground (blue low to yellow high).}
    \label{fig:Lidar-capture}
\end{figure}

\textbf{{UAV RGB imagery.} 
}
For the full set of plots, we collected overlapping RGB imagery using a DJI Matrice 4E (SZ DJI Technology Co., Ltd., Shenzhen, China) in two campaigns. We captured imagery with high overlap and mixed viewing geometry (including oblique views at 70 and 90\,m altitude) to support SfM/MVS reconstruction and increase the chance of observing trunk surfaces through gaps in the canopy and understory. To~improve absolute positioning consistency across flights, we corrected the onboard UAV GNSS trajectory via post-processed kinematic (PPK) processing using reference data from a nearby CORS station (ARMO), and~we used the corrected solution to update image geotags before reconstruction. An~example UAV RGB flight trajectory is shown in Figure~\ref{fig:Lidar-capture}a.

\textbf{{UAV LiDAR.}}
We acquired UAV laser scanning (ULS) data with an Inertial Labs RESEPI payload (Inertial Labs, Inc., Paeonian Springs, VA, USA) integrating a Hesai XT-32 LiDAR (Hesai Technology, Shanghai, China).
The XT-32 is a 360\(^\circ\) mid-range scanner with a 0.05--120\,m ranging capability and up to 640k\,pts/s in single-return mode, with~a typical range precision of 0.5\,cm (1$\sigma$) and a ranging accuracy of $\pm$1\,cm~\cite{HesaiXT32Spec,InertialLabsRESEPIXT32}. Figure~\ref{fig:Lidar-capture}b summarizes the LiDAR flight paths. The~LiDAR survey used dense, back-and-forth flight lines (a lawnmower grid) to provide uniform coverage and reduce directional bias in canopy and stem observations. We flew the UAV at ~50 m above ground level, yielding a point density of {$\approx 1407$ pts/m$^{2}$.} 

The~resulting georeferenced point clouds serve as a baseline geometry source for trunk isolation and DBH~fitting.

\begin{figure}[H]

    \includegraphics[width=\linewidth]{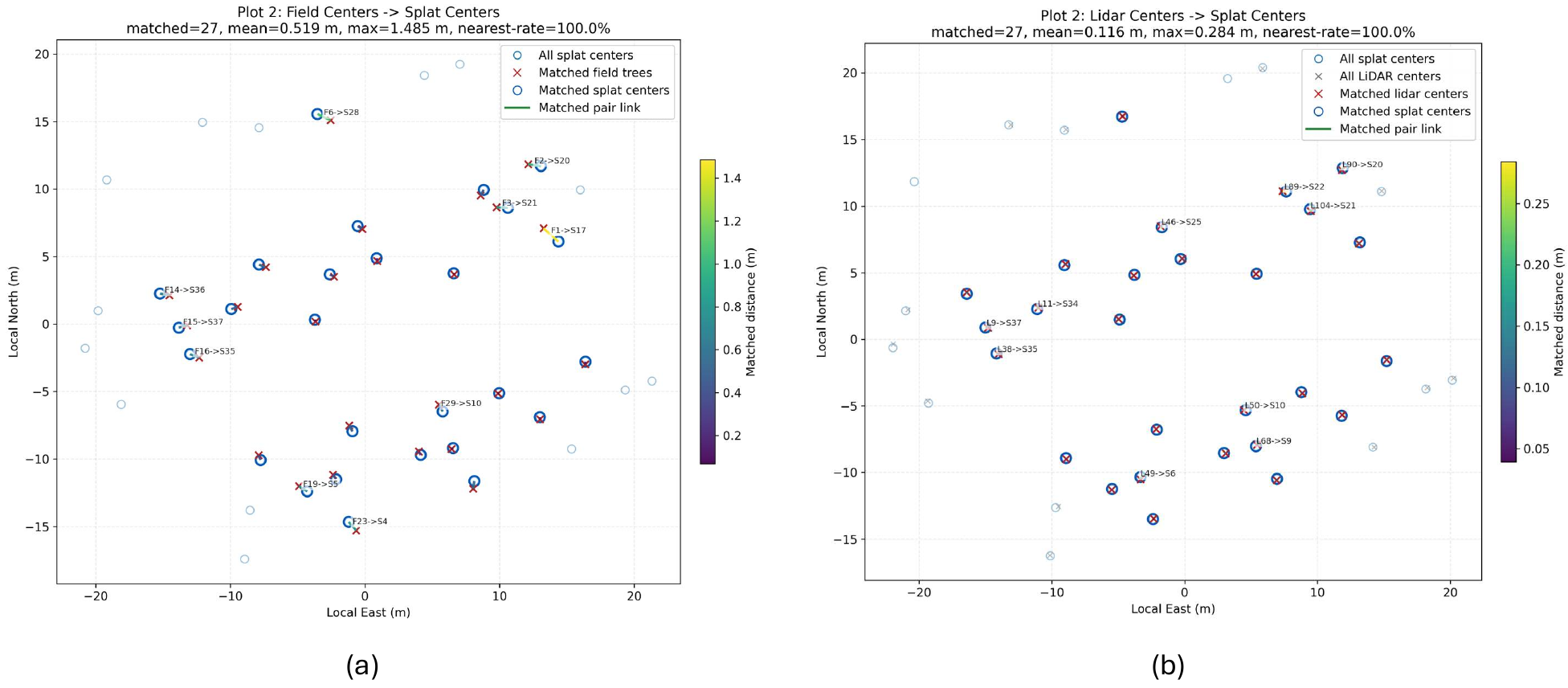}
    \caption{{\textbf{{Example of field-to-reconstruction stem matching.}} Plot-level overlay illustrating the manual pairing between field-inventoried pine stems and reconstructed trunk instances (after removing small non-pine trees). Lines connect matched pairs. The~mean planimetric offset is 0.776\,m for field-to-TreeDGS matches (210 pairs across 10 plots) and 0.270\,m for LiDAR-to-TreeDGS matches. These offsets reflect combined field distance/bearing uncertainty and georegistration error, and~they are used only as a sanity check after {visual verification.} }}
    \label{fig:Matching_Example}
\end{figure}

{\textbf{{Pairing workflow and quality control.}} Across all plots, we measured 458 stems in the field. To~create the per-tree evaluation set, we paired each field-inventoried pine stem to a reconstructed trunk instance using a standardized, spatially driven workflow. For~each plot, we converted the field-measured (bearing and distance) from the surveyed plot center into georeferenced $(x,y)$ coordinates. For~the reconstructed data (TreeDGS and LiDAR), we represented each trunk instance by the centroid of its segmented trunk points near breast height. First, we generated candidate associations using nearest-neighbor matching within the plot boundary, and~then we manually verified each pair in a plot-level visualization that overlays the field points, reconstructed instances, and~high-resolution aerial context (orthomosaic/point cloud; Figure~\ref{fig:Matching_Example}). Two authors performed pairing; uncertain or ambiguous cases (e.g., clustered stems, partially reconstructed trunks, or~missing instances) were discussed and, if~unresolved, excluded from the paired set. This resulted in 210 pair stems. During~fieldwork, we also recorded a plot walkthrough video, which served as an additional reference to confirm tree neighborhoods when needed. Table~\ref{tab:plot_summary} reports the per-plot pine stem counts and DBH distribution statistics.}

\section{Proposed~Pipeline}
\label{sec:pipeline}

\subsection{Problem~Statement}
\label{sec:problem}

Given a set of $N$ high-resolution UAV RGB images $\{I_i\}_{i=1}^{N}$, our goal is to estimate the diameter at breast height (DBH) for each tree instance in the scene.
DBH is defined as the trunk diameter at a fixed height above local ground, and~we follow the field protocol used in this study ($h_{\mathrm{BH}}\in[1.37,1.40]$\,m).

\begin{figure*}
    \centering
    \includegraphics[width=0.8\linewidth]{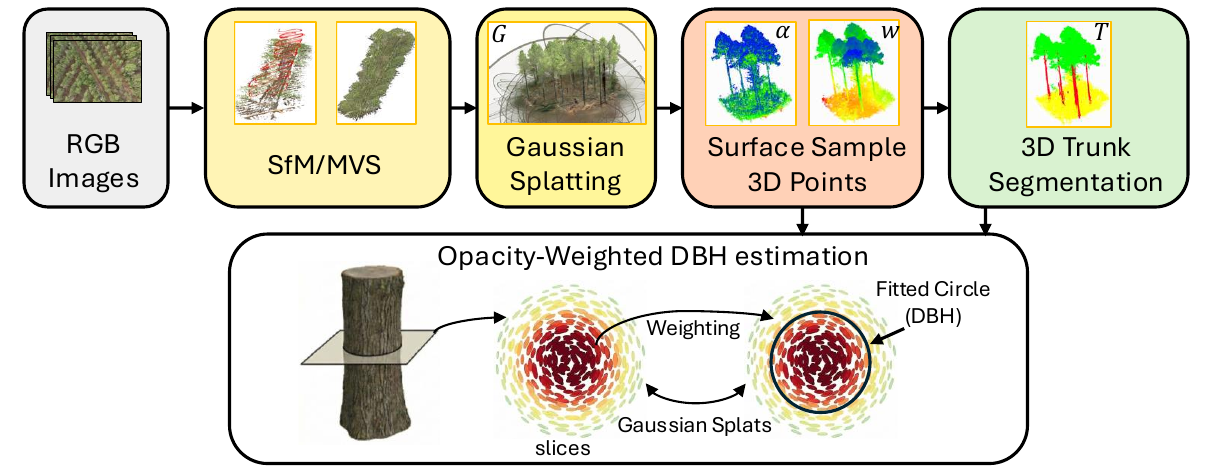}
    \caption{\textbf{{TreeDGS pipeline.}} RGB images are reconstructed with SfM/MVS and represented as 3D Gaussian splats $G$. We surface-sample dense points using opacity $\alpha$ and weights $w$, segment trunks $T$, and~estimate DBH by opacity-weighted circle fitting on multiple {trunk~slices.} Inset colors: $\alpha$ and $w$ are visualized with a low-to-high colormap (cool colors = low values; warm colors = high values); in $T$, trunk points are red and non-trunk points are green/yellow.}
    \label{TreeDGS_pipeline}
\end{figure*}

Our approach is geometry-driven: we reconstruct camera poses and an initial 3D structure with SfM, densify it with OpenMVS~\cite{openmvs_github}, optimize a Gaussian splatting~\cite{kerbl20233d, zhang2024rade} scene representation, and~then measure DBH from trunk-only geometry using opacity-aware sampling and robust fitting (Figure~\ref{TreeDGS_pipeline}): 

\begin{equation}
\label{eq:pipeline_overview}
\{I_i\}
\rightarrow (\{\mathbf{P}_i\}, \mathcal{X}_{\text{dense}})
\rightarrow \mathcal{G}
\rightarrow \{\mathcal{T}_t\}
\rightarrow \{\widehat{\mathrm{DBH}}_t\},
\end{equation}
{where} 
 $\mathbf{P}_i$ are the calibrated camera matrices, $\mathcal{X}_{\text{dense}}$ is a densified point set from OpenMVS, $\mathcal{G}$ is the optimized Gaussian field, and~$\mathcal{T}_t$ denotes trunk points for tree instance $t$.
\mbox{Sections~\ref{sec:sfm_openmvs}--\ref{sec:dbh_weighted}} detail the SfM/OpenMVS reconstruction, Gaussian optimization, trunk extraction through 3D segmentation~\cite{xiang2025forestformer3d} with opacity cues, and~the final DBH fitting~procedure.

\subsection{Structure-from-Motion and Multi-View~Stereo}
\label{sec:sfm_openmvs}

We estimate camera poses and an initial scene structure using a Structure-from-Motion (SfM) pipeline~\cite{pan2024glomap} adapted to high-resolution UAV forest imagery, where repeated textures and partial occlusions can reduce matching reliability. 
To improve correspondence quality, we use a customized TopicFM~\cite{giang2024topicfm+} model trained on a mixture of (i) MegaDepth~\cite{li2018megadepth}, (ii)~100 synthetic UAV nadir forest scenes, and~(iii) 200 available real 3D models from the Coolant~Dataset.

\textbf{{Coarse-to-fine matching.}}
Because our UAV imagery is high resolution, we perform matching in a coarse-to-fine manner~\cite{leroy2024grounding}:
we first obtain coarse correspondences on downsampled images to establish robust global alignment, and we then refine matches locally at higher resolution.
This reduces computation while maintaining accurate pixel localization needed for stable pose~estimation.

\textbf{{SfM reconstruction.}}
SfM takes the matched correspondences across overlapping image pairs and estimates the following:
(i) camera poses $\{\mathbf{P}_i\}$ and (ii) a sparse 3D point set.
We use standard robust estimation and bundle adjustment from GLOMAP~\cite{pan2024glomap} to refine the reconstruction.
We keep notation minimal and denote the resulting calibrated cameras as:
\begin{equation}
\mathbf{P}_i = \mathbf{K}_i [\mathbf{R}_i \mid \mathbf{t}_i].
\end{equation}
{These} 
 cameras are used both for Gaussian training (Section~\ref{sec:radegs}) and for the multi-view opacity tests during surface sampling (Section~\ref{sec:opacity_sampling}).

\textbf{{Coarse densification with OpenMVS~\cite{openmvs_github}.}}
After SfM, we densify geometry using OpenMVS.
{OpenMVS produces depth information and a denser point set $\mathcal{X}_{\text{dense}}$, which provides stronger geometric support than sparse SfM points in many well-observed regions (e.g., ground, canopy, and~partially visible trunk segments).}
{We note that $\mathcal{X}_{\text{dense}}$ is not assumed to be complete or coherent in the breast-height band. In~our setting, MVS trunk coverage can be sparse and fragmented due to occlusion and weak texture. 
}
{Instead, we use $\mathcal{X}_{\text{dense}}$ only as a coarse prior to initialize the Gaussian centers for faster and more stable optimization, following common 3DGS practice~\cite{kerbl20233d,jung2024rain,wu2025sparse2dgs}.}

\subsection{Reconstruction with Gaussian~Splats}
\label{sec:radegs}

Given calibrated cameras $\{\mathbf{P}_i\}$ and an MVS-densified initialization
$\mathcal{X}_{\text{dense}}$ from \mbox{Section~\ref{sec:sfm_openmvs},} we optimize a
3D Gaussian Splatting (3DGS) scene representation.
We adopt RaDe-GS~\cite{zhang2024rade}, which builds on the real-time 3DGS
renderer~\cite{kerbl20233d} while explicitly rasterizing depth in a way
that is also useful for our subsequent surface-consistent point~sampling.

\textbf{{Gaussian parameterization.}}
The scene is represented as a set of anisotropic 3D Gaussians
$G_k=(\boldsymbol{\mu}_k,\mathbf{s}_k,\mathbf{R}_k,\alpha_k,\mathbf{c}_k)$,
where $\boldsymbol{\mu}_k\in\mathbb{R}^3$ is the Gaussian center,
$\mathbf{s}_k\in\mathbb{R}^3_{+}$ are axis-aligned scales in the local frame,
$\mathbf{R}_k\in SO(3)$ is the per-Gaussian orientation, $\alpha_k\in[0,1]$ is
the opacity, and~$\mathbf{c}_k$ denotes appearance parameters (e.g.,
spherical-harmonic features as in~\cite{kerbl20233d}). We initialize
$\{\boldsymbol{\mu}_k\}$ from $\mathcal{X}_{\text{dense}}$ and optimize all
parameters following RaDe-GS~\cite{zhang2024rade}. {Note that RaDe-GS interleaves photometric optimization with density control (splitting and pruning). Therefore, the~Gaussian set can be progressively densified and refined over time, improving coverage in regions that are only weakly supported by the MVS initialization.}

\textbf{{RaDe-GS depth-plane formulation.}}
A key difference from vanilla 3DGS is that RaDe-GS associates each Gaussian and
view with a local ray-distance plane in screen space.
For a view $v$, let $\mathbf{u}_{k,v}\in\mathbb{R}^2$ be the projected mean of
Gaussian $k$, and~let $t_{k,v}$ be its ray-distance (range) in that view
(i.e., $\|\mathbf{R}_v\boldsymbol{\mu}_k+\mathbf{t}_v\|_2$ under perspective
projection). RaDe-GS additionally provides a 2D slope
$\mathbf{g}_{k,v}\in\mathbb{R}^2$, so that the Gaussian's predicted ray-distance at a nearby pixel
$\mathbf{u}$ is approximated by
\begin{equation}
\label{eq:radegs_depth_plane}
d_{k,v}(\mathbf{u})
\;=\;
t_{k,v} \;+\; \mathbf{g}_{k,v}^{\top}(\mathbf{u}_{k,v}-\mathbf{u}).
\end{equation}
We use this depth-plane formulation in Section~\ref{sec:opacity_sampling} to
define a depth-aware, surface-consistent opacity integral at arbitrary 3D query
points.

\textbf{{Adaptive training with RLGS.}}
We integrate RLGS~\cite{li2025rlgs} as an online controller that
adapts selected training hyper-parameters based on observed optimization
dynamics. In~our setting, this is primarily used to stabilize opacity and
geometry quality for the downstream DBH pipeline.
After convergence, the~optimized Gaussian field is denoted by $\mathcal{G}$ and
serves as the source representation for dense point~sampling.

\subsection{Surface Sampling with~Opacity}
\label{sec:opacity_sampling}

\begin{figure}[H]

    \includegraphics[width=\linewidth]{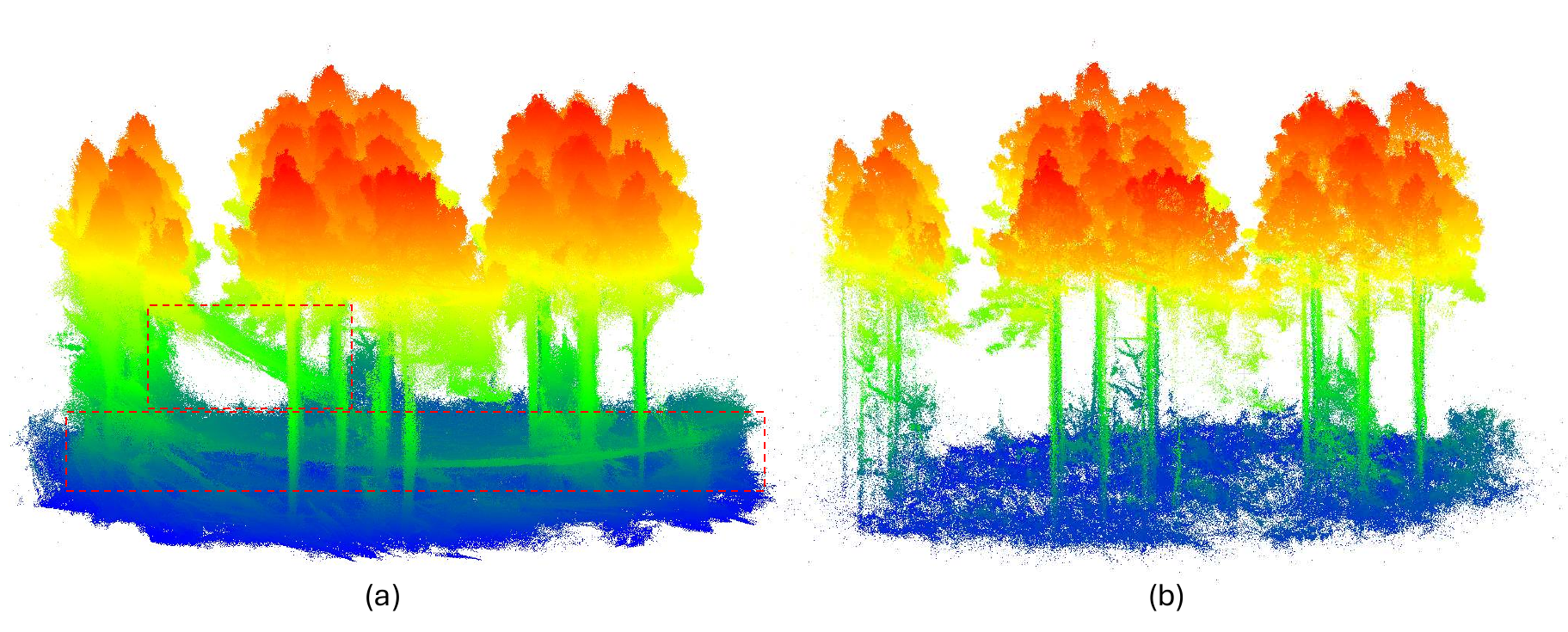}
    
    \caption{\textbf{{Comparison of back-projection points with multi-view depth consistency and the proposed sample~points.}} (\textbf{a}) Depth-fusion baseline: rendered depth points are back-projected into 3D and kept only if they pass multi-view depth-consistency checks. This produces large sheet-like artifacts near the ground/understory and incomplete stem surfaces (red dashed boxes), despite appearing consistent across views. (\textbf{b}) Our opacity-guided surface sampling with depth-aware point-wise compositing give a stem geometry suitable for {DBH estimation. Blue indicates lower elevations near ground, and red/orange indicates higher canopy points.}}
    \label{fig:depthconsistencysample_vs_proposedsample}
\end{figure}

Our downstream stages (3D trunk segmentation and DBH fitting) require a point cloud
that is (i) dense enough to represent thin stems at stand-off distance and
(ii) geometrically trustworthy under heavy occlusion from foliage and understory.
This is not automatic from a 3D Gaussian Splatting (3DGS) reconstruction: the optimized
field $\mathcal{G}$ is a collection of overlapping volumetric primitives whose parameters
are trained through front-to-back alpha compositing rather than an explicit surface loss
~\cite{kerbl20233d,zhang2024rade}. As~a result, two seemingly straightforward exports are
unreliable in our UAV forest regime.
First, exporting one point per Gaussian (the means) produces a fragmented and extremely sparse cloud
 
 which is insufficient for robust 3D segmentation and
measurement. Second, back-projecting rendered depth maps into 3D and fusing them via multi-view
depth-consistency preserves view-consistent sheet-like artifacts from semi-transparent vegetation while eroding already-scarce trunk evidence,
especially when stems are only a few pixels wide or intermittently visible (see Figure~\ref{fig:depthconsistencysample_vs_proposedsample}).
Therefore, we sample densely in 3D, but~we accept and score samples using tests that are consistent with
the same compositing model that governs 3DGS rendering and~optimization.

{\textbf{{Opacity-guided stochastic sampling.}}} Our goal is to convert the optimized Gaussian field $\mathcal{G}$ into a point cloud that is dense enough for trunk segmentation and DBH fitting.
For each Gaussian
$G_i=(\boldsymbol{\mu}_i,\mathbf{s}_i,\mathbf{R}_i,\alpha_i)$, we draw candidate
offsets $\boldsymbol{\xi}_{ij}\sim\mathcal{N}(\mathbf{0},\mathbf{I}_3)$ in the
local frame and map them to world coordinates. To~focus sampling on reliable
(high-opacity) regions, we apply Bernoulli thinning controlled by
$\alpha_i$:
\begin{equation}
\label{eq:opacity_sampling}
\begin{aligned}
\mathbf{p}_{ij} &= \boldsymbol{\mu}_i
\;+\; \mathbf{R}_i\left(\mathbf{s}_i \odot \boldsymbol{\xi}_{ij}\right), \\
b_{ij} &\sim \mathrm{Bernoulli}(\alpha_i).
\end{aligned}
\end{equation}
We keep $\mathbf{p}_{ij}$ if $b_{ij}=1$, in expectation each Gaussian contributes
$\alpha_i M$ points when drawing $M$ candidates. {Here, $M$ is the maximum number of candidate samples drawn per Gaussian (before opacity-guided thinning) and directly controls the point density and sampling cost. In~all experiments we use $M=100$ (i.e., at~most 100 candidate draws per splat). Since candidates are retained with probability $\alpha_i$, the~expected number of retained samples per Gaussian is $\alpha_i M$, which keeps the computational budget bounded while still allocating higher sample density to high-opacity (more reliable) regions.}

{\textbf{{Front-to-back compositing and accumulated alpha.}}} In 3DGS, solidity is not defined by an explicit surface. It is defined implicitly by how Gaussians accumulate opacity under the renderer’s front-to-back alpha compositing. Therefore, any surface sampling rule that aims
to match what the model optimized should be expressed using the same compositing mechanism.
For a view $v$ and pixel $\mathbf{u}$, the~renderer composites Gaussians
front-to-back using transmittance tracking. If~$\alpha_{k,v}(\mathbf{u})$ is the
per-Gaussian alpha contribution at $\mathbf{u}$, we define $T_0(\mathbf{u})=1$
and update
\begin{equation}
\label{eq:alpha_compositing}
\begin{aligned}
w_{k,v}(\mathbf{u})
&= \alpha_{k,v}(\mathbf{u})\,T_{k-1}(\mathbf{u}), \\
T_k(\mathbf{u})
&= T_{k-1}(\mathbf{u})\bigl(1-\alpha_{k,v}(\mathbf{u})\bigr).
\end{aligned}
\end{equation}
so the accumulated alpha mask is
$A_v(\mathbf{u})=\sum_k w_{k,v}(\mathbf{u}) = 1 - T_{\text{final}}(\mathbf{u})$.
In our code this mask is read from the renderer's alpha channel and used as a
visibility gate for sampled~points.

The per-pixel accumulated alpha mask $A_v(\mathbf{u})$ tells us whether a ray intersects the reconstructed mass somewhere along the ray, but~it does not tell us whether a particular 3D sample $\mathbf{p}$ lies on the visible surface, behind~it, or~in front of it (all can project to the same pixel). For~surface extraction we therefore need a depth-aware test that measures how much opacity accumulates in front of $\mathbf{p}$ along the viewing ray, using the same compositing model that governs training. RaDe-GS provides exactly this query by integrating point-wise alphas with a depth-plane~formulation.

{\textbf{{Depth-aware point-wise opacity integration.}}} Beyond~per-pixel alpha, RaDe-GS provides an integration query at an
arbitrary 3D point $\mathbf{p}$: it returns the projected pixel coordinate
$\mathbf{u}_v(\mathbf{p})$ and a cumulative opacity value
$\tilde{\alpha}_v(\mathbf{p})$ that measures how much opacity is accumulated
in front of $\mathbf{p}$ along the viewing ray.
Let $r_v(\mathbf{p})=\|\mathbf{R}_v\mathbf{p}+\mathbf{t}_v\|_2$ denote the point
ray-distance in view $v$.
For a Gaussian $k$, RaDe-GS evaluates a ray-space quadratic form using a
precomputed inverse covariance 
over the 3D offset
\begin{equation}
\label{eq:radegs_point_offset}
\Delta\mathbf{u}_{k,v}(\mathbf{p})
=
\begin{bmatrix}
\mathbf{u}_{k,v}-\mathbf{u}_v(\mathbf{p}) \\
t_{k,v} - \min\!\bigl(r_v(\mathbf{p}),\, d_{k,v}(\mathbf{u}_v(\mathbf{p}))\bigr)
\end{bmatrix},
\end{equation}
and converts it to an alpha contribution via an exponential falloff 

\begin{equation}
\alpha_{k,v}(\mathbf{p})
\;\propto\;
\alpha_k \exp\!\left(
-\tfrac{1}{2}\Delta\mathbf{u}_{k,v}(\mathbf{p})^{\top}
\boldsymbol{\Sigma}_{k,v}^{-1}
\Delta\mathbf{u}_{k,v}(\mathbf{p})
\right).
\end{equation}
The $\min(\cdot)$ depth clamping in
Equation~\eqref{eq:radegs_point_offset} is critical for surface consistency:
once $\mathbf{p}$ lies behind the locally rasterized surface depth
$d_{k,v}(\cdot)$, the~depth residual term saturates rather than increases with
$r_v(\mathbf{p})$, preventing spurious ``volumetric'' accumulation behind the
surface. Then, RaDe-GS composites these point-wise alphas using the same
front-to-back transmittance update as in Equation~\eqref{eq:alpha_compositing},
yielding the cumulative opacity
\begin{equation}
\label{eq:pointwise_opacity_integral}
\begin{aligned}
\tilde{\alpha}_v(\mathbf{p})
&=
\sum_k \alpha_{k,v}(\mathbf{p})\,T_{k-1,v}(\mathbf{p}), \\
T_{k,v}(\mathbf{p})
&=
T_{k-1,v}(\mathbf{p})\bigl(1-\alpha_{k,v}(\mathbf{p})\bigr).
\end{aligned}
\end{equation}

{\textbf{{Multi-view surface consistency and reliability scores.}}} Since DBH fitting is highly sensitive to even small amounts of geometric contamination, we aggregate evidence across the calibrated views and attach a per-point reliability that downstream steps can threshold or
use as a weight. Concretely, we (i) only allow a view to contribute if the point
projects to a foreground pixel in that view, and~we (ii) summarize the point-wise
opacity integrals across contributing views conservatively.
Let $\mathcal{P}=\{\mathbf{p}_{ij}\mid b_{ij}=1\}$ be the candidate samples
retained after the opacity-guided thinning in Equation~\eqref{eq:opacity_sampling}.
For each $\mathbf{p}\in\mathcal{P}$, we evaluate its support over the calibrated
views; a view $v$ contributes only if $\mathbf{p}$ projects inside the image and
lands on a non-background pixel in the rendered accumulated alpha mask $A_v$.
Specifically, with~$\mathbf{u}_v(\mathbf{p})$ and 
the projection of $\mathbf{p}$,
we define
\begin{equation}
{m_v(\mathbf{p})
=
\mathbf{1}\!\left(A_v(\mathbf{u}_v(\mathbf{p}))>0\right).}
\end{equation}

Then, we store two per-point reliability signals:
\begin{equation}
\label{eq:opacity_reliability}
\bar{\alpha}(\mathbf{p})
=
\min_{v:\,m_v(\mathbf{p})=1}\tilde{\alpha}_v(\mathbf{p}),
\qquad
w(\mathbf{p})=\sum_v m_v(\mathbf{p}).
\end{equation}

Here, $\bar{\alpha}(\mathbf{p})$ is a conservative multi-view opacity estimate, {and $w(\mathbf{p})$ counts the number of views in which $\mathbf{p}$ projects onto a foreground pixel, i.e.,~$A_v(\mathbf{u}_v(\mathbf{p}))>0$.} In practice, for~surface extraction, we keep points with $\bar{\alpha}(\mathbf{p})>\tau$ and $w(\mathbf{p})>0$.
{Especially for this work, we set $\tau=0$ to keep all points while storing
$(\bar{\alpha},w)$ for downstream weighting.}

\subsection{Semantic Trunk~Extraction}
\label{sec:trunk_seg}

The sampled point cloud from Section~\ref{sec:opacity_sampling} contains trunks mixed with foliage, understory vegetation, and~occasional floating samples caused by Gaussians with nonzero opacity in free space. To~isolate stem geometry for DBH measurement, we used ForestFormer3D~\cite{xiang2025forestformer3d}, a~3D semantic segmentation model, to~predict a class label for each sampled point and to assign per-tree instance IDs when available. In~our pipeline, the~per-point opacity reliability $\bar{\alpha}$ is kept as an auxiliary output feature alongside $(x,y,z)$. We retain only points predicted as trunk to form a trunk-only cloud for each tree instance $t$, denoted by $\mathcal{T}_t=\{(\mathbf{x}_k,\bar{\alpha}_k)\}$. This separation step reduces contamination from branches and understory, while the retained opacity values enable reliability-weighted fitting downstream. The~trunk-only instances are then passed to the geometric measurement stage described~next.

\subsection{Opacity-Weighted DBH~Measurement}
\label{sec:dbh_weighted}

Given a trunk-only point set for tree instance $t$,
\begin{equation}
\begin{aligned}
\mathcal{T}_t &= \left\{(\mathbf{x}_k,\bar{\alpha}_k)\right\}_{k=1}^{N_t}, \\
\mathbf{x}_k &= (x_k,y_k,z_k)^\top \in \mathbb{R}^3,\quad
\bar{\alpha}_k \in [0,1].
\end{aligned}
\end{equation}

our goal is to estimate the diameter at breast height (DBH), i.e.,~the trunk diameter at $h_{\mathrm{BH}}\approx 1.3\text{--}1.4$~m above local ground. A~key challenge is that (i) the sampled points are volumetric (many points lie inside the trunk cross-section, not only on the boundary), and~(ii) residual non-trunk points and floaters can persist even after segmentation. We address both issues using an opacity-weighted solid-circle RANSAC in each horizontal slice, followed by a robust height-wise taper~regression.

DBH is defined relative to local ground height. For~each tree instance, we estimate a ground elevation $z_g$ by querying a digital terrain model (DTM) at the tree location (e.g., nearest-neighbor lookup at the mean trunk $(x,y)$), and~we convert all points to height above ground:
\begin{equation}
z_g = \mathrm{DTM}\!\left(\tfrac{1}{N_t}\sum_{k=1}^{N_t}(x_k,y_k)\right),
\qquad
h_k = z_k - z_g.
\end{equation}
We perform all subsequent slicing and DBH evaluation in the $(x,y)$ plane as a function of $h$.

Next, we construct a sequence of slice centers $\{h_s\}$ with spacing $\Delta z$ and thickness $H$ (a slab of height $H$), and~we collect 2D points in each slab:
\begin{equation}
\label{eq:slicing}
\begin{aligned}
h_s &= s\,\Delta z,\quad s=0,1,\dots, \\
\mathcal{S}_s &= \left\{(x_k,y_k,\bar{\alpha}_k)\;\middle|\;
\left|h_k-h_s\right|\le \tfrac{H}{2}\right\}.
\end{aligned}
\end{equation}

Slices with fewer than a small minimum number of points {$n_{\min}$} are discarded. This step produces a set of candidate diameters along the lower~stem.

Unlike classical TLS stem fitting where points often lie near the circumference, our sampling in Section~\ref{sec:opacity_sampling} draws points from Gaussian volumes; therefore, many slice points legitimately fall inside the trunk cross-section. For~this reason we treat inliers as belonging to a filled disk rather than a narrow ring around a circle. This choice makes the estimator consistent with volumetric sampling and substantially more stable under partial~visibility.

\textbf{{Opacity-weighted solid-circle RANSAC in each slice.}}
For slice $s$, we have 2D points $\{(\mathbf{q}_k,w_k)\}_{k=1}^{n}$ with $\mathbf{q}_k=(x_k,y_k)^\top$ and opacity weights $w_k=\bar{\alpha}_k$.
Each RANSAC hypothesis performs the following {steps:} 

(i) Weighted sampling: Draw three distinct indices $(k_1,k_2,k_3)$ without replacement with $\Pr(k)\propto w_k$.

(ii) Circle fit from three points: Fit the circle through the sampled points using the algebraic form
$x^2+y^2+a x+b y+d=0$.
We solve for $(a,b,d)$ after subtracting the slice mean (for numerical stability), and we then recover
$\mathbf{c}=(-\tfrac{a}{2},-\tfrac{b}{2})^\top$ and $r^2=\lVert\mathbf{c}\rVert_2^2-d$ (adding the mean back to $\mathbf{c}$).

(iii) Solid inliers and validity checks: A point is an inlier if it lies inside the disk:
$\lVert \mathbf{q}_k-\mathbf{c}\rVert_2^2 \le r^2$.
We discard hypotheses whose radius $r$ is abnormally large or that yield fewer than $\rho n$ inliers.

(iv) Opacity-weighted scoring: For the remaining hypotheses, we score
\begin{equation}
\label{eq:weighted_score}
\begin{aligned}
S(\mathbf{c},r)
&=
\frac{\sum_{k=1}^{n} w_k\,
\mathbf{1}\!\left(\lVert \mathbf{q}_k-\mathbf{c}\rVert_2^2 \le r^2\right)}{r^{p}}, \\
&\quad p>0.
\end{aligned}
\end{equation}
and select $(\hat{\mathbf{c}}_s,\hat{r}_s)=\arg\max_{\mathbf{c},r} S(\mathbf{c},r)$, reporting $\hat{d}_s=2\hat{r}_s$.

\textbf{{Height-wise taper regression and DBH prediction.}}
Even after per-slice robust fitting, some slices can be corrupted by branch attachments, residual foliage, or~incomplete sampling. Therefore, we fit a simple taper model on the lower stem using RANSAC with a negative-slope prior:
\begin{equation}
\label{eq:taper}
\hat{d}(h)=\beta_0 + \beta_1 h,
\qquad
{-\kappa \le \beta_1 < 0},
\end{equation}
where RANSAC discards outlier slices {using an absolute residual threshold $\epsilon$ on diameter ($\left|\hat{d}_s-(\beta_0+\beta_1 h_s)\right|\le \epsilon$)} and the slope constraint enforces physically plausible taper (diameter should not increase with height over short stem segments). {The additional bound $\kappa$ prevents unrealistically steep taper caused by noisy slice diameters.} 
 In practice, we fit Equation~\eqref{eq:taper} over progressively larger height ranges starting near the ground (to avoid upper-canopy contamination) until a stable inlier set is found. Finally, we report DBH as:
\begin{equation}
\label{eq:dbh_final}
\mathrm{DBH} = \hat{d}(h_{\mathrm{BH}}),
\qquad h_{\mathrm{BH}}\in[1.37,1.40]\;\text{m} \,.
\end{equation}

\textbf{{Role of opacity in robustness.}}
The opacity weights $w_k=\bar{\alpha}_k$ originate from the multi-view rendering consistency test in Section~\ref{sec:opacity_sampling}. Points that are only weakly supported (e.g., floaters, thin vegetation, or~ambiguous geometry) tend to have small $\bar{\alpha}$ and therefore (i) are sampled less often during hypothesis generation and (ii) contribute little to the inlier score in Equation~\eqref{eq:weighted_score}. This coupling between rendering consistency and geometric fitting is the central mechanism behind our opacity-weighted measurement~mode.

\section{Experiments}
\label{sec:experiments}

\subsection{Experimental~Settings}
We evaluate TreeDGS on the 10 circular field plots described in Section~\ref{sec:materials}.
Field DBH was measured with a diameter tape at breast height ($h_{\mathrm{BH}}=1.37$\,m above ground, Section~\ref{sec:materials}).
We compare direct DBH measurement under the following reconstruction sources and fitting strategies:
(i) UAV LiDAR: DBH estimation from the aerial LiDAR point cloud using (a) cylinder fitting~\cite{malladi2025digiforests} and (b) our slice-wise circle fitting (non-weighted and intensity-weighted variants).
(ii) TreeDGS: DBH estimation from the point cloud sampled from the optimized 3D Gaussian field (Section~\ref{sec:opacity_sampling}), using the same cylinder and circle fitting variants, with~opacity-based weighting available only for TreeDGS.
To ensure a fair comparison, trunk/instance segmentation for both LiDAR and TreeDGS inputs is performed using the same ForestFormer3D checkpoint~\cite{xiang2025forestformer3d}. {If ForestFormer3D misses a matched tree instance (e.g., due to strong bending/occlusion) or if the downstream fitting cannot return a DBH estimate, we count that tree as a failure when computing the Success Rate (SR) in Table~\ref{tab:dbh_comparison_all_plots}.}

\textbf{{Cylinder fitting baseline.}}
We follow DigiForests~\cite{malladi2025digiforests} for cylinder fitting.
A cylinder is fit to each trunk point cloud using a RANSAC-initialized model.
Each RANSAC round samples five points, estimates the cylinder center and axis from the covariance SVD, and~keeps models whose axis is within $\pi/8$ of vertical and radius $\le 0.5\,\mathrm{m}$.
The RANSAC cylinder is then refined by least squares, updating the cylinder center, axis, and~radius to minimize a robust (Geman--McClure) sum of squared point-to-cylinder distance residuals.
Iterations stop when the parameter update norm falls below $\lVert \Delta x \rVert < 10^{-5}$, where \mbox{$\Delta x \in \mathbb{R}^7$} stacks the center shift, rotation/axis update, and~radius change.
If the least-squares refinement increases the inlier ratio, it replaces the RANSAC estimate; otherwise, the RANSAC cylinder is retained.
Finally, the~diameter at breast height is reported as $\mathrm{DBH}=2r$ from the selected cylinder~radius.

\textbf{{Circle fitting variants.}}
For the proposed circle fitting, we evaluate the following two reliability modes:
non-weighted (nw), which treats all points equally, and~weighted (w), which uses a per-point reliability score as a weight (Equation~\eqref{eq:opacity_reliability}) in both hypothesis sampling and inlier scoring (Equation~\eqref{eq:weighted_score}).
For TreeDGS, this reliability is the multi-view opacity consistency estimated by our renderer.
For LiDAR, we additionally report an intensity-weighted analogue that uses normalized return intensity as a proxy weight.
{In our UAV data, intensity-weighting helps reduce the strong positive bias of non-weighted solid-circle fitting (Table~\ref{tab:dbh_comparison_all_plots}), but~it still underperforms LiDAR cylinder fitting and has lower SR, consistent with raw airborne LiDAR intensity being speckled and unstable without careful radiometric correction and calibration (e.g., sensitivity to range, incidence angle, and~scanner settings)~\cite{Hofle2007IntensityCorrection,Kashani2015RadiometricProcessing,Kaasalainen2011ALSCalibration,Wu2021IncidenceAngleCorrection,Yan2018IntensityBanding}.}

\begin{figure}
    \includegraphics[width=0.95\linewidth]{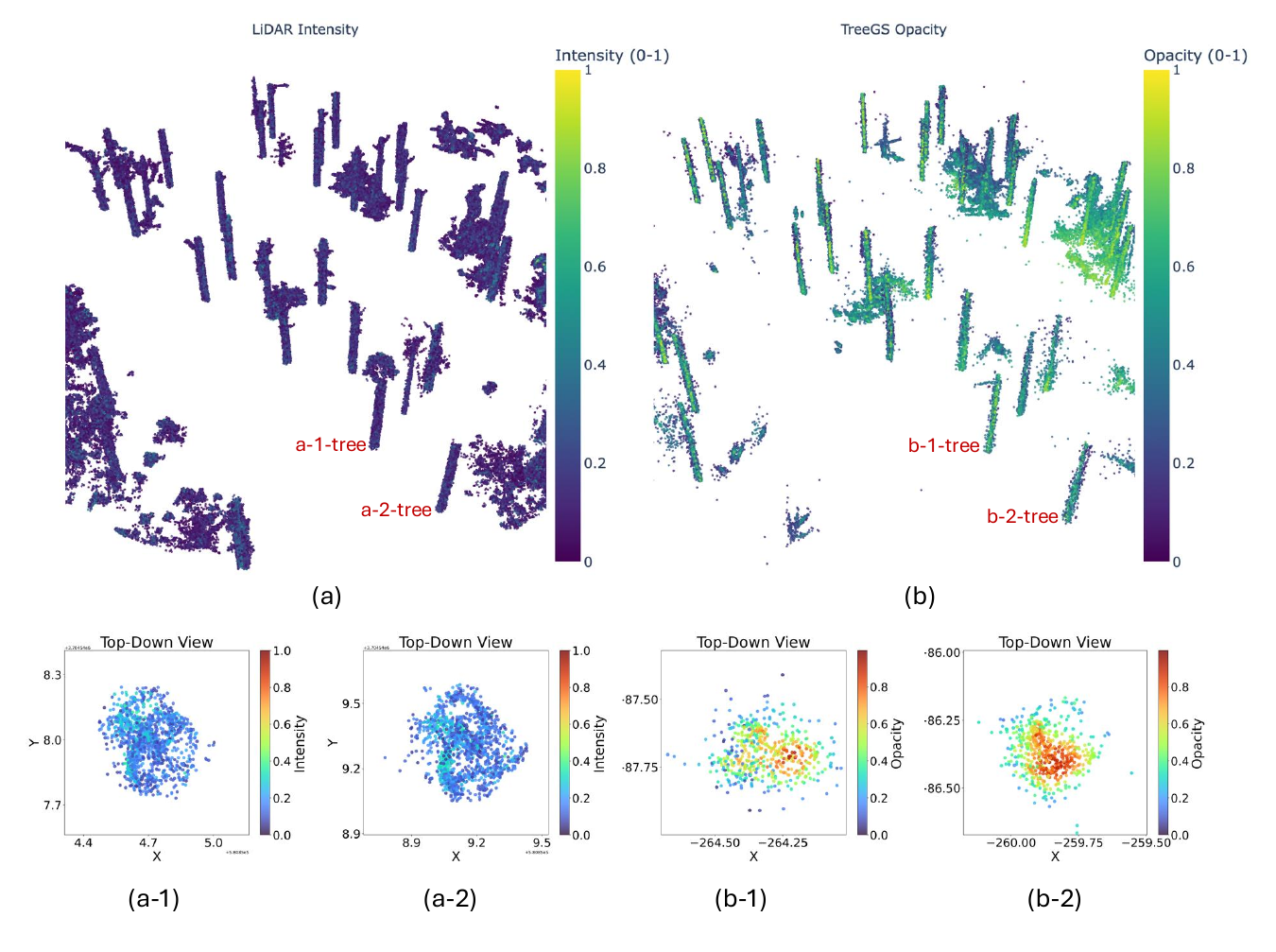}
    \caption{\textbf{{LiDAR intensity vs. TreeDGS opacity as a reliability signal at a same 5m slice height (Plot~2).}} (\textbf{a}) UAV LiDAR point cloud colored by normalized return intensity. (\textbf{b}) TreeDGS surface samples colored by opacity (0--1). Bottom: zoomed top-down views of two example stems ((\textbf{a-1},\textbf{a-2}) intensity; (\textbf{b-1},\textbf{b-2}) opacity). Compared to LiDAR intensity, the~opacity values form a clearer high-confidence core around the stem cross-sections. This provides per-point confidence cue for DBH circle fitting. LiDAR has 281.4K points and TreeDGS has {66.6K points.}}
    \label{fig:opacity_vs_intensity}
\end{figure}

\begin{figure}
    \includegraphics[width=\linewidth]{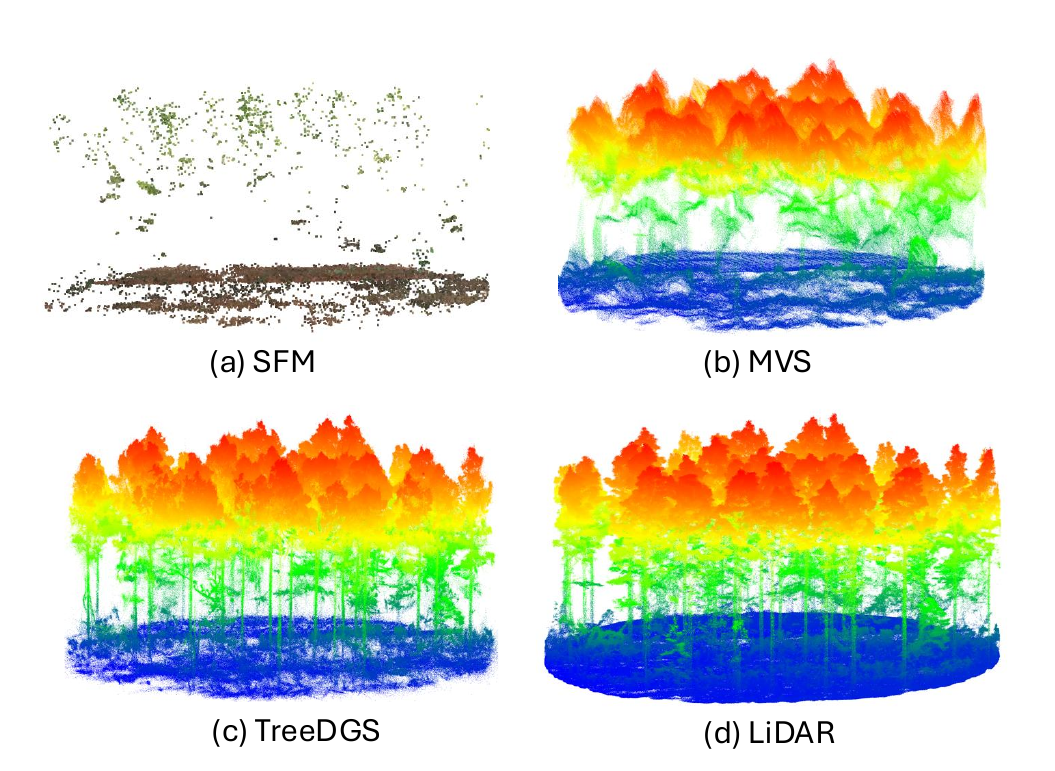}
    \caption{\textbf{{Qualitative comparison of point density and trunk completeness (Plot~1).}} (\textbf{a}) SfM points are sparse with limited stem coverage. (\textbf{b}) OpenMVS densification improves density but remains incomplete on trunks under canopy occlusion. (\textbf{c}) TreeDGS extracts a dense point set from the optimized Gaussians via opacity-guided sampling {($M=100$)} and RaDe-GS depth-aware cumulative-opacity tagging, yielding more continuous trunk support near breast height. (\textbf{d}) UAV LiDAR provides strong structure but can still be stem-sparse/contaminated depending on scan geometry and occlusion. Colors indicate height {above ground. }}
    \label{fig:pc_comparison}
\end{figure}

Figure~\ref{fig:opacity_vs_intensity} visualizes this gap: opacity yields a cleaner, geometry-aligned confidence core around stem cross-sections compared to noisy intensity. In~addition, we omitted SfM/MVS benchmarking because breast-height trunk points were insufficient for stable segmentation and fitting in this dataset ({Figure~\ref{fig:pc_comparison}).}

\begin{figure}[H]

    \includegraphics[width=\linewidth]{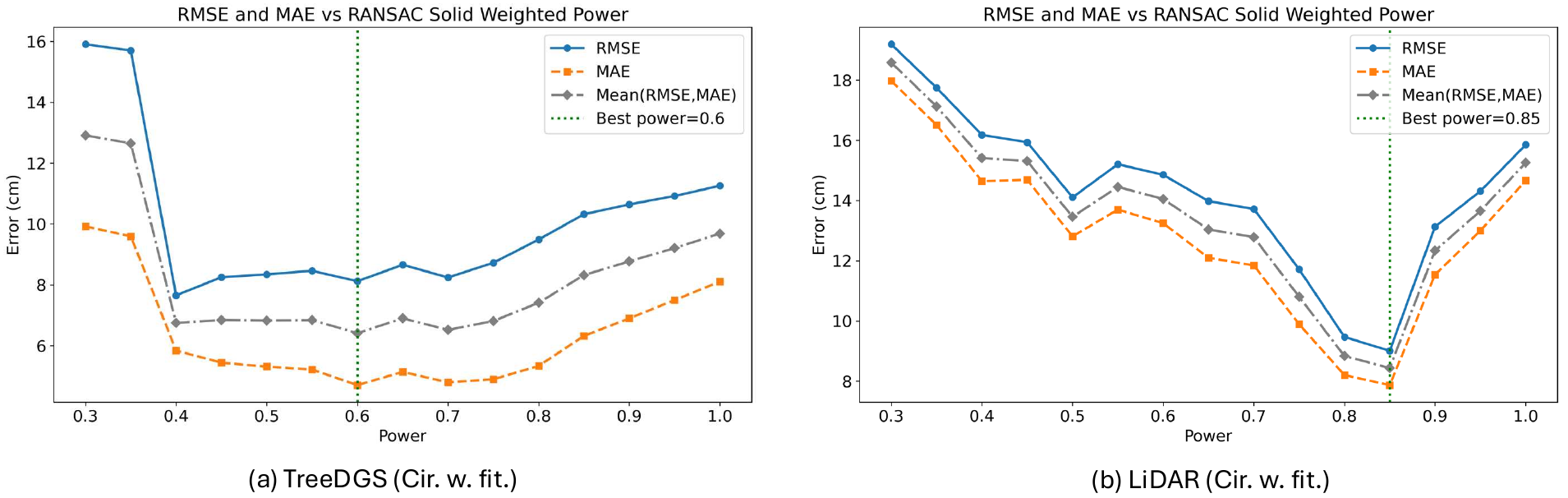}
    \caption{{\textbf{{Held-out validation for selecting the radius penalty exponent $p$ in Equation~\eqref{eq:weighted_score}.}} We sweep $p$ for weighted solid-circle RANSAC and measure RMSE/MAE on the held-out validation split. (\textbf{a}) TreeDGS opacity-weighted circle fitting selects $p=0.6$. (\textbf{b}) LiDAR intensity-weighted circle fitting selects $p=0.85$. We keep these values fixed when reporting all test results in Table~\ref{tab:dbh_comparison_all_plots}.}}
    \label{fig:power_search}
\end{figure}

Across all circle-fitting variants, we use slice thickness $H=1.0$\,m and slice spacing $\Delta z=0.1$\,m, discarding slices with fewer than $5$ points.
For each slice, we run solid-circle RANSAC (Equation~\eqref{eq:weighted_score}) with $K=2000$ hypotheses, a~minimum inlier fraction $\rho_{\min}=0.1$, and~radius bounds $r_{\min}=0.02$\,m and $r_{\max}=1.0$\,m.
{We select the radius regularization exponent $p$ in Equation~\eqref{eq:weighted_score} via held-out validation on a random selection of 10\% of the matched stems and then keep it fixed for all test experiments. We use the same validation/test split IDs for all compared methods for fair comparison. Concretely, from~the 210 matched trees in Table~\ref{tab:plot_summary}, we randomly reserve 21 trees (10\%) for validation and report all numbers in Table~\ref{tab:dbh_comparison_all_plots} on the remaining disjoint test set (N=189). This is why the per-plot denominators in Table~\ref{tab:dbh_comparison_all_plots} differ from Table~\ref{tab:plot_summary}. We obtain $p=0.6$ for TreeDGS opacity weighting and $p=0.85$ for the LiDAR intensity-weighted analogue (Figure~\ref{fig:power_search}).}

The taper model in Equation~\eqref{eq:taper} uses RANSAC with residual threshold $\epsilon=2$\,cm (max trials $T=1000$, min samples $3$), requiring at least $10$ inlier slices and enforcing $\beta_1<0$.
Finally, DBH is reported as $\mathrm{DBH}=\hat d(h_{\mathrm{BH}})$ (Equation~\eqref{eq:dbh_final}) evaluated at $h_{\mathrm{BH}}=1.37$\,m.

\vspace{-15pt}

We report four standard error measures between estimated and field DBH:
root mean squared error (RMSE), relative RMSE (RRMSE, normalized by the mean field DBH), mean absolute error (MAE), and~mean error (ME, bias).
All errors are reported in centimeters, and~lower is better for RMSE/RRMSE/MAE while ME closer to 0 indicates lower systematic bias. {We additionally report the Success Rate (SR) as the fraction of trees in the test split for which a method returns a valid DBH estimate (Table~\ref{tab:dbh_comparison_all_plots}). RMSE/RRMSE/MAE/ME are computed over the successful subset, while SR captures failures (e.g., missing instances or insufficient inlier slices for the taper RANSAC).}

\begin{table}[t]
\centering
\scriptsize
\setlength{\tabcolsep}{1.8pt}
\renewcommand{\arraystretch}{0.94}
\caption{\textbf{Per-plot DBH accuracy and robustness (All--5).} RMSE/MAE/ME are in cm, RRMSE is in \%, and SR is success rate. Abbrev.: L/T = LiDAR/TreeDGS; Cyl = cylinder; CN/CW = circle non-weighted/weighted.}
\label{tab:dbh_comparison_all_plots}
\begin{tabular}{@{}c l r r r r c@{}}
\toprule
\textbf{Plot} & \textbf{Meth.} & \textbf{RMSE} & \textbf{RRMSE} & \textbf{MAE} & \textbf{ME} & \textbf{SR} \\
\midrule
\multirow{6}{*}{All} & L-Cyl & 7.66 & 28.09 & 5.23 & 1.50 & 189/189 \\
 & L-CN & 16.61 & 60.96 & 14.81 & 14.67 & 187/189 \\
 & L-CW & 10.29 & 36.97 & 8.79 & 1.02 & 176/189 \\
 & T-Cyl & 13.29 & 48.69 & 10.25 & $-$6.46 & 189/189 \\
 & T-CN & 9.61 & 35.22 & 6.97 & 5.79 & 189/189 \\
 & T-CW & \textbf{4.79} & \textbf{17.54} & \textbf{3.70} & \textbf{$-$0.38} & 189/189 \\
\midrule
\multirow{6}{*}{1} & L-Cyl & 8.16 & 26.14 & 5.71 & \textbf{$-$0.04} & 19/19 \\
 & L-CN & 10.12 & 32.42 & 8.57 & 7.57 & 19/19 \\
 & L-CW & 9.94 & 32.04 & 8.04 & 2.01 & 18/19 \\
 & T-Cyl & 11.69 & 37.47 & 10.99 & $-$10.99 & 19/19 \\
 & T-CN & 7.58 & 24.29 & 5.91 & 2.63 & 19/19 \\
 & T-CW & \textbf{4.94} & \textbf{15.83} & \textbf{3.72} & $-$2.04 & 19/19 \\
\midrule
\multirow{6}{*}{2} & L-Cyl & 5.36 & 18.84 & 4.34 & \textbf{0.38} & 22/22 \\
 & L-CN & 13.27 & 46.66 & 11.92 & 11.92 & 22/22 \\
 & L-CW & 8.60 & 30.25 & 7.12 & $-$0.59 & 22/22 \\
 & T-Cyl & 13.43 & 47.21 & 9.84 & $-$5.49 & 22/22 \\
 & T-CN & 8.43 & 29.65 & 7.26 & 7.09 & 22/22 \\
 & T-CW & \textbf{5.07} & \textbf{17.82} & \textbf{3.18} & 1.88 & 22/22 \\
\midrule
\multirow{6}{*}{3} & L-Cyl & 12.92 & 49.55 & 7.75 & 4.47 & 20/20 \\
 & L-CN & 20.89 & 80.11 & 18.69 & 18.69 & 20/20 \\
 & L-CW & 11.85 & 45.44 & 10.32 & 3.55 & 20/20 \\
 & T-Cyl & 10.49 & 40.23 & 8.77 & $-$7.35 & 20/20 \\
 & T-CN & 16.39 & 62.83 & 9.47 & 8.15 & 20/20 \\
 & T-CW & \textbf{5.93} & \textbf{22.73} & \textbf{4.35} & \textbf{$-$0.10} & 20/20 \\
\midrule
\multirow{6}{*}{4} & L-Cyl & 6.63 & 27.06 & 5.81 & 2.94 & 16/16 \\
 & L-CN & 19.34 & 78.87 & 17.95 & 17.95 & 16/16 \\
 & L-CW & 9.59 & 35.77 & 7.78 & $-$1.41 & 12/16 \\
 & T-Cyl & 25.24 & 102.93 & 16.57 & 3.84 & 16/16 \\
 & T-CN & 11.46 & 46.73 & 7.59 & 4.95 & 16/16 \\
 & T-CW & \textbf{5.74} & \textbf{23.40} & \textbf{4.53} & \textbf{$-$0.62} & 16/16 \\
\midrule
\multirow{6}{*}{5} & L-Cyl & 6.20 & 21.87 & 4.27 & 2.37 & 21/21 \\
 & L-CN & 17.30 & 60.96 & 16.60 & 16.60 & 20/21 \\
 & L-CW & 11.61 & 40.92 & 10.99 & 3.26 & 20/21 \\
 & T-Cyl & 12.39 & 43.69 & 11.01 & $-$7.98 & 21/21 \\
 & T-CN & 5.84 & 20.60 & 4.64 & 3.57 & 21/21 \\
 & T-CW & \textbf{4.38} & \textbf{15.46} & \textbf{4.01} & \textbf{$-$2.10} & 21/21 \\
\bottomrule
\end{tabular}
\end{table}

\begin{table}[t]
\ContinuedFloat
\centering
\scriptsize
\setlength{\tabcolsep}{1.8pt}
\renewcommand{\arraystretch}{0.94}
\caption{\textbf{Per-plot DBH accuracy and robustness (cont., plots 6--10).}}
\begin{tabular}{@{}c l r r r r c@{}}
\toprule
\textbf{Plot} & \textbf{Meth.} & \textbf{RMSE} & \textbf{RRMSE} & \textbf{MAE} & \textbf{ME} & \textbf{SR} \\
\midrule
\multirow{6}{*}{6} & L-Cyl & 4.96 & 18.41 & 3.27 & 1.27 & 22/22 \\
 & L-CN & 14.03 & 52.15 & 12.86 & 12.86 & 22/22 \\
 & L-CW & 9.45 & 35.12 & 7.60 & \textbf{0.26} & 22/22 \\
 & T-Cyl & 8.64 & 32.10 & 7.91 & $-$7.45 & 22/22 \\
 & T-CN & 8.91 & 33.10 & 8.23 & 8.23 & 22/22 \\
 & T-CW & \textbf{3.15} & \textbf{11.72} & \textbf{2.48} & 0.91 & 22/22 \\
\midrule
\multirow{6}{*}{7} & L-Cyl & 9.16 & 44.08 & 7.23 & 1.92 & 17/17 \\
 & L-CN & 15.79 & 75.98 & 14.86 & 14.55 & 17/17 \\
 & L-CW & 7.79 & 33.96 & 6.31 & $-$2.32 & 13/17 \\
 & T-Cyl & 13.66 & 65.77 & 8.15 & \textbf{$-$0.28} & 17/17 \\
 & T-CN & 12.01 & 57.81 & 10.57 & 9.68 & 17/17 \\
 & T-CW & \textbf{5.91} & \textbf{28.44} & \textbf{5.03} & 3.27 & 17/17 \\
\midrule
\multirow{6}{*}{8} & L-Cyl & 9.65 & 33.11 & 7.80 & 2.61 & 16/16 \\
 & L-CN & 20.61 & 70.70 & 18.21 & 18.21 & 16/16 \\
 & L-CW & 11.07 & 36.67 & 9.77 & \textbf{1.11} & 15/16 \\
 & T-Cyl & 11.83 & 40.59 & 10.97 & $-$7.64 & 16/16 \\
 & T-CN & 5.42 & 18.59 & 3.96 & 3.83 & 16/16 \\
 & T-CW & \textbf{3.68} & \textbf{12.63} & \textbf{3.14} & $-$1.69 & 16/16 \\
\midrule
\multirow{6}{*}{9} & L-Cyl & 6.18 & 23.13 & 4.20 & \textbf{$-$0.38} & 16/16 \\
 & L-CN & 16.27 & 62.05 & 15.09 & 15.09 & 15/16 \\
 & L-CW & 11.08 & 42.26 & 10.33 & 1.92 & 15/16 \\
 & T-Cyl & 8.73 & 32.66 & 7.73 & $-$7.73 & 16/16 \\
 & T-CN & 8.50 & 31.81 & 6.85 & 5.17 & 16/16 \\
 & T-CW & \textbf{3.93} & \textbf{14.72} & \textbf{3.08} & $-$1.26 & 16/16 \\
\midrule
\multirow{6}{*}{10} & L-Cyl & \textbf{3.79} & \textbf{12.84} & \textbf{3.00} & \textbf{$-$0.30} & 20/20 \\
 & L-CN & 16.95 & 57.41 & 14.93 & 14.80 & 20/20 \\
 & L-CW & 10.73 & 35.95 & 9.23 & 0.84 & 19/20 \\
 & T-Cyl & 12.07 & 40.88 & 11.40 & $-$11.15 & 20/20 \\
 & T-CN & 6.28 & 21.27 & 5.15 & 4.10 & 20/20 \\
 & T-CW & 4.42 & 14.95 & 3.73 & $-$2.35 & 20/20 \\
\bottomrule
\end{tabular}
\end{table}

\subsection{Results}
\label{subsec:quant_dbh}

Table~\ref{tab:dbh_comparison_all_plots} summarizes per-plot performance and includes a pooled `All' row.
{All results in Table~\ref{tab:dbh_comparison_all_plots} are reported on the disjoint test split (N~=~189) after reserving 21 of the 210~matched trees (10\%) for validation/hyperparameter selection (see Experimental settings).}
TreeDGS with opacity-weighted circle fitting achieves the best overall accuracy (pooled \mbox{RMSE~$=4.79$\,cm;} MAE $=3.70$\,cm) with negligible bias (ME $=-0.38$\,cm), while maintaining a high success rate (189/189).
{Compared with the UAV LiDAR cylinder-fitting baseline (pooled RMSE $=7.66$\,cm), TreeDGS reduces pooled RMSE by $\approx 37\%$ while using only \mbox{RGB imagery.}}

Ablations also show that the fitting model must match the reconstructed geometry.
Applying a classical cylinder fit~\cite{malladi2025digiforests} to points sampled from TreeDGS performs poorly (13.29\,cm RMSE; ME $=-6.46$\,cm), consistent with our discussion in Section~\ref{sec:dbh_weighted}: TreeDGS sampling produces slice-wise cross-sections rather than a thin ring of surface points.
Similarly, unweighted circle fitting is substantially less accurate (9.61\,cm RMSE), showing that down-weighting low-confidence samples using multi-view opacity is a better approach.
{On LiDAR, the~same trend holds: non-weighted solid-circle fitting strongly overestimates \mbox{(ME $=14.67$\,cm)}, while intensity-weighted fitting with a radius penalty substantially reduces this bias \mbox{(ME $=1.02$\,cm)}, though~it still underperforms LiDAR cylinder fitting.}

{TreeDGS (Cir.\ w.\ fit.) attains the lowest RMSE in nine of ten plots (all except Plot~10) and achieves the best pooled performance (All row). In~Plot~2 it slightly outperforms the LiDAR cylinder baseline (5.07 vs.\ 5.36\,cm RMSE). The~largest gap appears in Plot~3, where LiDAR (cyl.\ fit.) reaches 12.92\,cm RMSE while TreeDGS (Cir.\ w.\ fit.) remains at 5.93\,cm. In~Plot~10, LiDAR (cyl.\ fit.) remains the best (3.79\,cm RMSE).}

{Although a solid-disk inlier model can in principle encourage larger radii when volumetric interior points dominate, in~our data the dominant failure mode for non-weighted fitting is positive bias driven by sparse exterior outliers (foliage/floaters) inflating the disk radius. The~weighted objective in Equation~\eqref{eq:weighted_score} counteracts this through (i)~reliability weights (opacity/intensity) and (ii) the explicit radius penalty $r^{-p}$, which requires proportionally more weighted inlier support to justify a larger radius. After~optimizing $p$ on held-out validation (Figure~\ref{fig:power_search}), the~estimator transitions from the positive-bias regime to near-unbiased behavior; the small pooled negative ME of $-0.38$\,cm for TreeDGS (Cir.\ w.\ fit.) indicates mild shrinkage when only a partial high-confidence arc of the trunk boundary is visible and is within typical measurement/registration noise.}

{Success Rate (SR) in Table~\ref{tab:dbh_comparison_all_plots} explicitly accounts for cases where a method cannot return a DBH estimate. For~example, LiDAR intensity-weighted circle fitting fails on 13 test trees (SR 176/189) because the taper RANSAC requires at least 10 inlier slices; intensity weighting can amplify slice-to-slice diameter variability and reduce the number of inlier slices below this threshold. In~contrast, the~non-weighted LiDAR circle fit fails on only two low-density trees (SR 187/189), suggesting that most additional failures are due to sensitivity to noisy intensity rather than fundamental point scarcity.}

\begin{figure}[H]
    \includegraphics[width=0.86\linewidth]{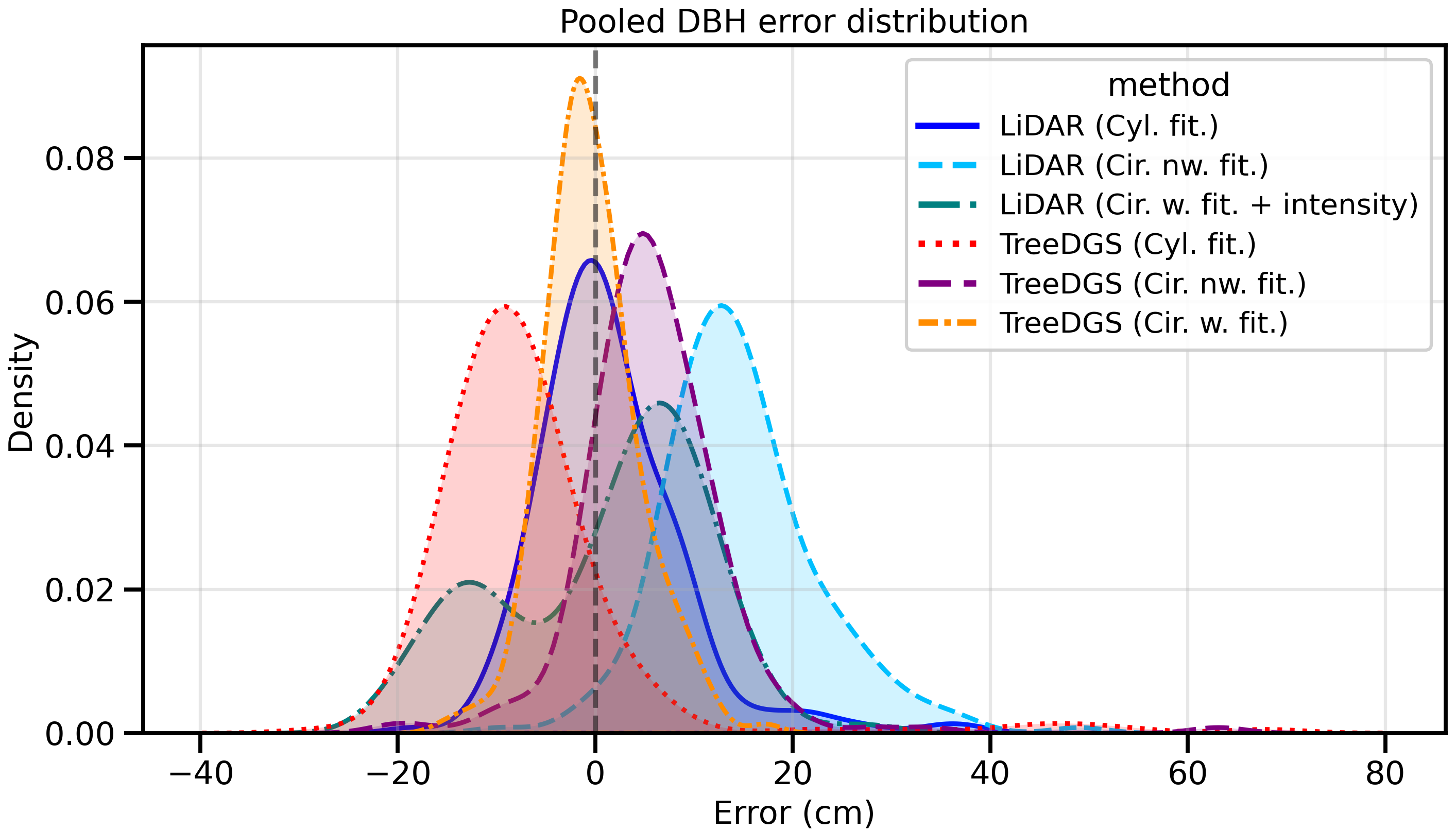}
\caption{{\textbf{{Aggregated DBH error~distributions.}} Kernel density estimates of signed per-tree DBH error (estimate$-$field, cm) pooled across all plots for {each method.} The vertical grey dashed line marks zero error (perfect agreement with field DBH). }}
    \label{fig:pooled_error_distribution}
\end{figure}

\begin{figure}
    \centering
    \includegraphics[width=0.86\linewidth]{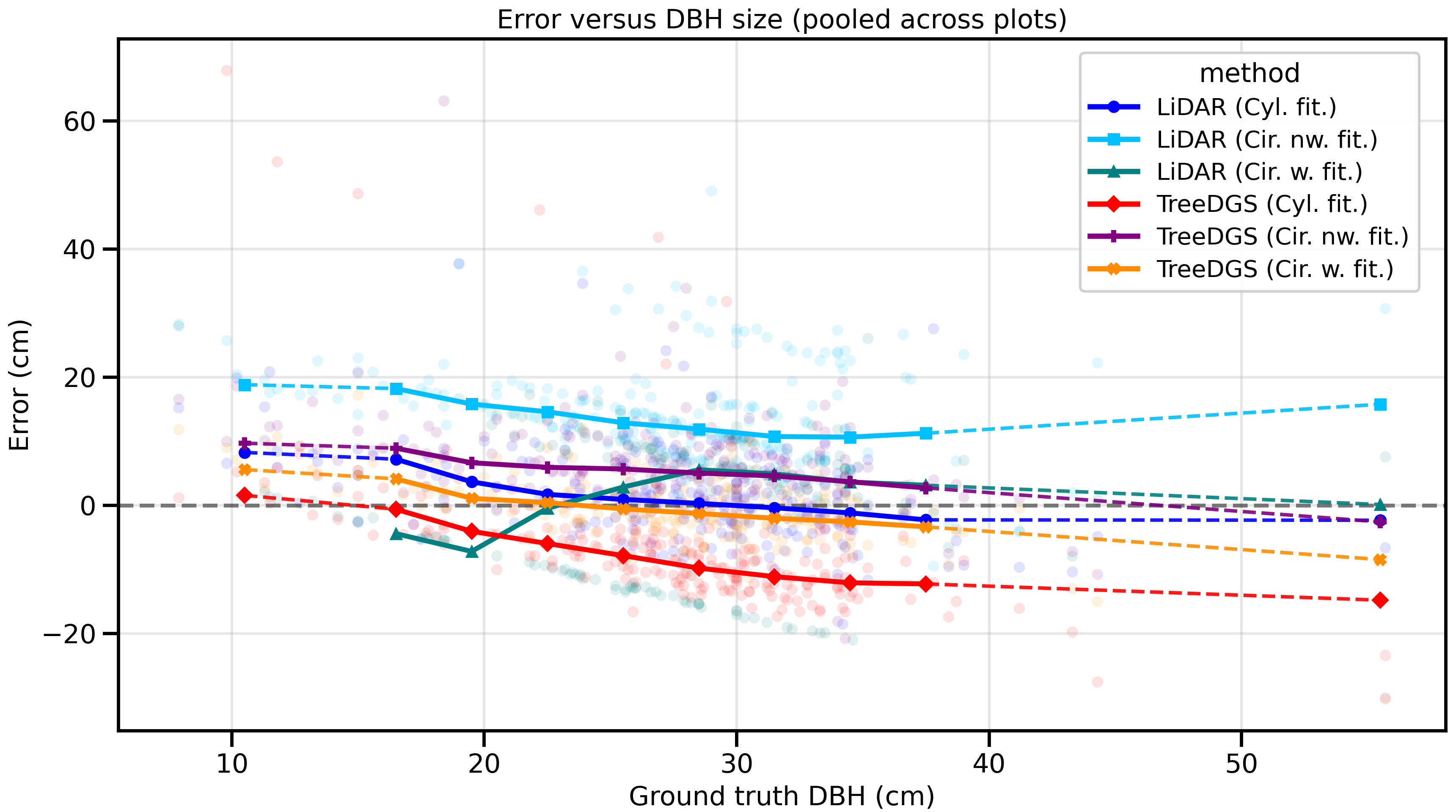}
\caption{{\textbf{{DBH error versus DBH size (aggregated across plots).}} Signed per-tree error (estimate-field, cm) versus ground-truth DBH on the test split, aggregated across all plots. Semi-transparent dots represent individual trees. For~each method, thick solid segments show a robust binned error trend. Dashed segments are only used to bridge DBH ranges with no observations, linking the nearest available bins without implying data exist {inside the gap.} }}
    \label{fig:error_vs_dbh}
\end{figure}

\begin{figure}
    \includegraphics[width=\linewidth]{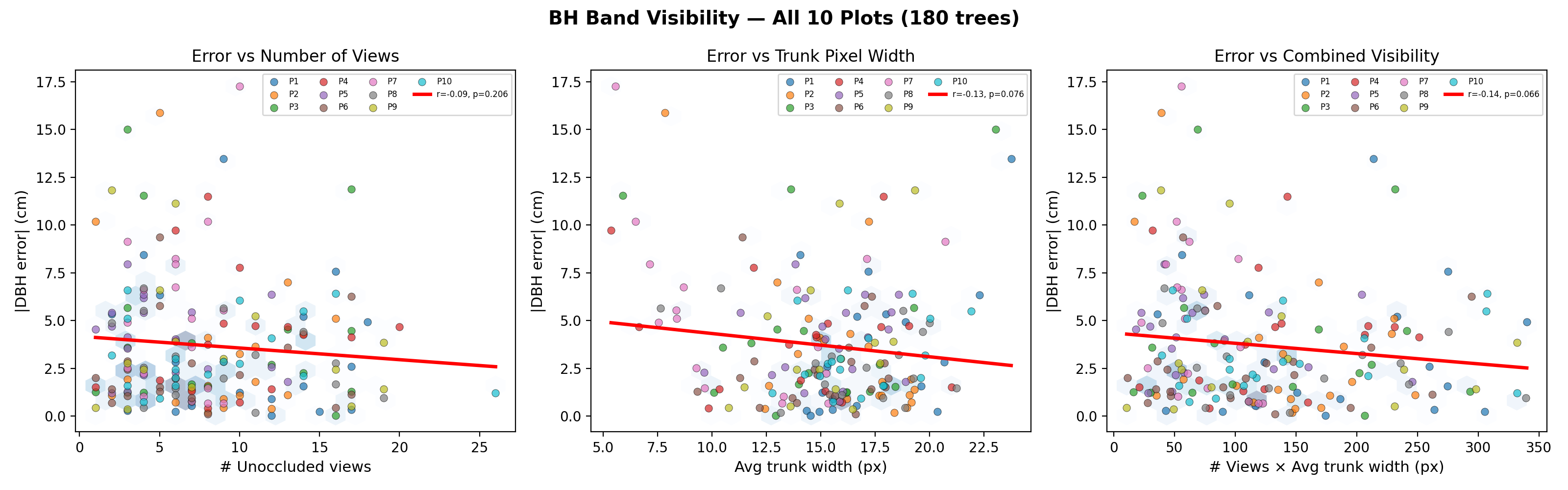}
\caption{{\textbf{{DBH error versus breast-height (BH) visibility in the input RGB~views.}} For each tree we compute (\textbf{left}) the number of unoccluded views in which the BH trunk band is visible, (\textbf{middle}) the average projected trunk width in pixels over those views, and~(\textbf{right}) a combined visibility score (\#views$\times$width). Each point is one tree from the test split (180 trees across 10 plots with at least one unoccluded BH view), colored by plot; blue hexbin shading indicates point density. The y-axis is the absolute DBH error for TreeDGS (Cir.\ w.\ fit.); red lines show least-squares trends (Pearson $r$ and $p$ in legend). Mean unoccluded views per tree $=8.2$ and mean trunk width $=15.4$\, {px.} }}
    \label{fig:bh_visibility_pooled_all_plots}
\end{figure}

\begin{figure}
    \centering
    \includegraphics[width=0.9\linewidth]{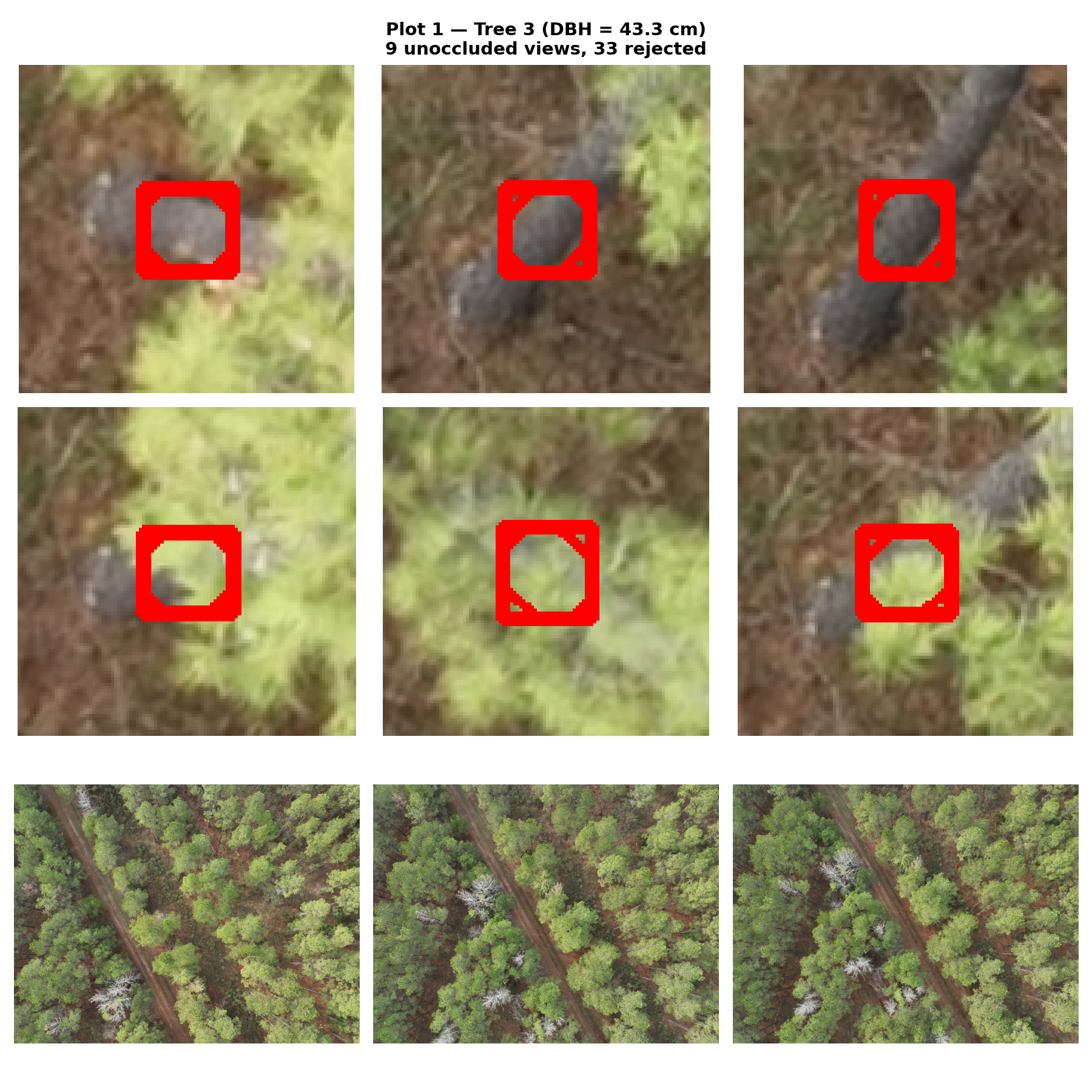}
\caption{{\textbf{{Example BH-band visibility classification used in Figure~\ref{fig:bh_visibility_pooled_all_plots}.}} We project a small trunk band around $h_{\mathrm{BH}}$ into each calibrated RGB view and classify the crop (red box) as unoccluded or occluded. (\textbf{Top}): views counted as unoccluded where bark is visible within the BH band. (\textbf{Middle}): views rejected as occluded because foliage covers the BH band. (\textbf{Bottom}): corresponding top-full-frame context, illustrating how small the BH band is relative to the canopy in stand-off UAV imagery.}}
    \label{fig:bh_visibility_example}
\end{figure}

{Beyond reporting pooled RMSE/MAE, we add three diagnostics to characterize robustness and failure modes. First, Figure~\ref{fig:pooled_error_distribution} plots pooled per-tree signed error distributions (estimate$-$field). The~opacity-weighted solid-circle variant yields the tightest distribution centered near zero, consistent with its lowest pooled RMSE/MAE and near-zero ME in Table~\ref{tab:dbh_comparison_all_plots}. In~contrast, the~non-weighted solid-circle fits show a pronounced positive shift and heavier right tails (radius inflation), while cylinder fitting on TreeDGS samples shifts negative (systematic underestimation). Second, Figure~\ref{fig:error_vs_dbh} reports signed error versus ground-truth DBH. Size-dependent bias is clearly visible for several baselines: TreeDGS (Cyl.\ fit.) increasingly underestimates as DBH grows (large negative mean error for the largest stems), and~LiDAR (Cir.\ nw.\ fit.) maintains strong positive bias across sizes. TreeDGS (Cir.\ w.\ fit.) shows substantially reduced size dependence: its binned mean error stays within single-digit centimeters and close to zero for the mid-range DBH values that dominate our plots (\(\sim\)18--38\,cm), while becoming more negative in the largest-DBH bin (which contains relatively few trees). Third, to~quantify the view-scarcity/occlusion bottleneck, Figure~\ref{fig:bh_visibility_pooled_all_plots} relates the absolute DBH error (TreeDGS (Cir.\ w.\ fit.)) to simple breast-height (BH) observability metrics computed from the calibrated RGB views: the number of unoccluded BH views, the~mean projected trunk width in pixels, and~their product. On~the 180/189 test trees with at least one unoccluded BH view, we observe weak but consistently negative correlations, indicating that higher BH visibility is associated with lower error. Importantly, the~largest outliers (\(>\)10\,cm) occur predominantly in the low-visibility regime (few unoccluded views and/or small projected trunk width), supporting our discussion that even with opacity reliability weighting, accurate DBH requires at least a few informative views around breast height. {Figure~\ref{fig:bh_visibility_example}} 
 shows an example of the BH-band view selection used for this analysis.}

\vspace{-12pt}

\newcommand{\mthead}[2]{\makecell[c]{\textbf{#1}\\{\scriptsize #2}}}

\begin{figure}
    \centering
    \includegraphics[width=0.9\linewidth]{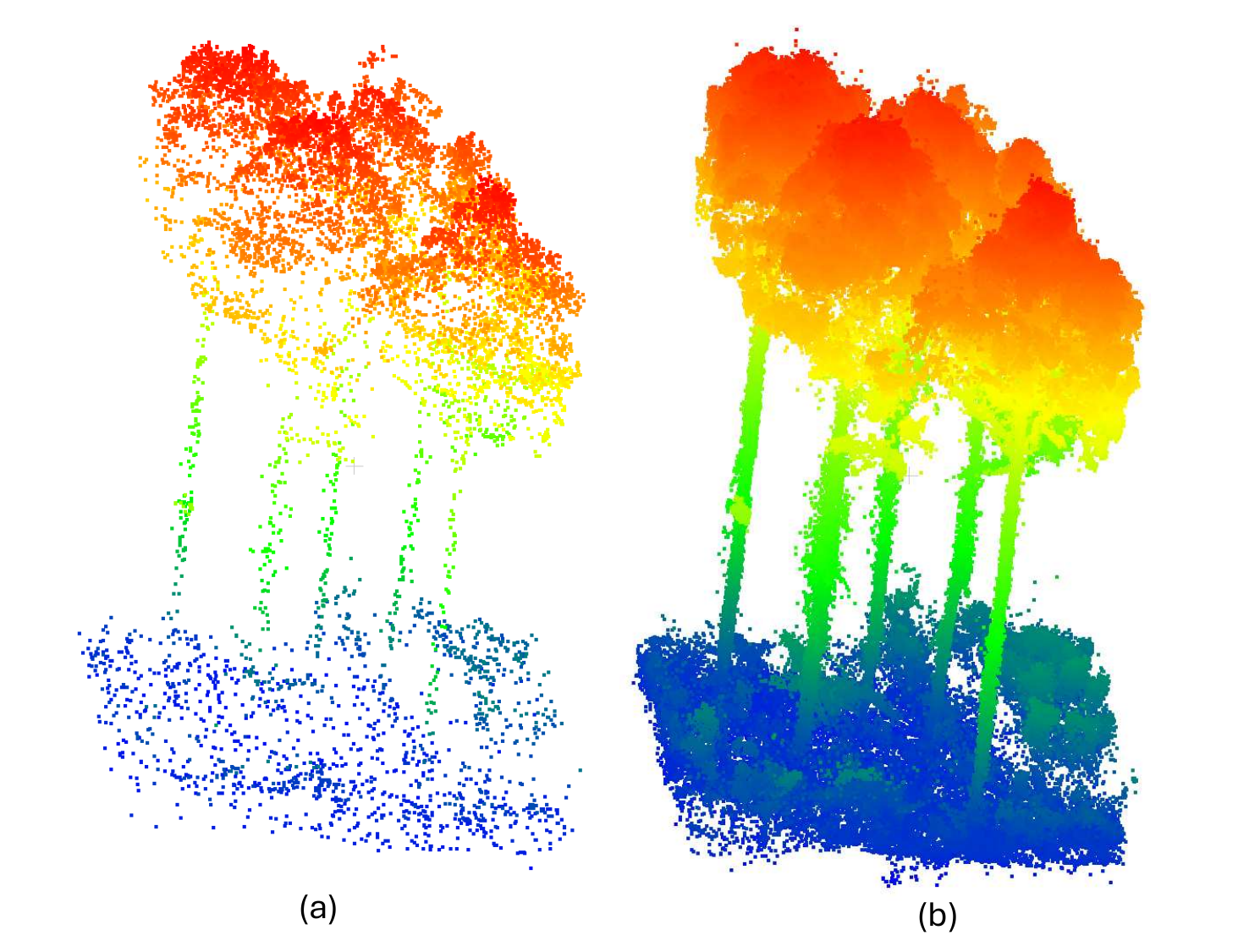}
    \caption{\textbf{{Mean-only vs. proposed point sampling from Gaussian Splats.}} (\textbf{a}) Exporting only Gaussian means (one point per splat) yields an extremely sparse, fragmented cloud that is insufficient for downstream tasks such as 3D segmentation. (\textbf{b}) Our opacity-guided surface sampling densifies geometry and preserves stem surfaces and {canopy structure.} Point colors encode height above ground (blue low to red/orange high).}
    \label{fig:Compare_mean_proposed_sample}
\end{figure}

We additionally report a simple baseline that exports one point per Gaussian by using the splat centers $\{\boldsymbol{\mu}_i\}$, similar to trunk-focused usage in prior 3DGS tree pipelines~\cite{Li2025Leafless}.
In our stand-off UAV forest scenes, this produces an extremely sparse and surface-incomplete point set ({Figure~\ref{fig:Compare_mean_proposed_sample}}), which breaks downstream instance segmentation with ForestFormer3D~\cite{xiang2025forestformer3d} and prevents reliable DBH fitting.
These failures highlight that surface-densified sampling is necessary in the stand-off UAV~setting.

\subsection{Ablation~Study}

\subsubsection{Sensitivity Analysis of Fitting~Parameters}
{In this section, we analyze the sensitivity of the DBH estimation stage to the main design parameters that govern slice construction (Equation~\eqref{eq:slicing}) and robust aggregation across height (Equation~\eqref{eq:taper}). We vary one parameter at a time, slice thickness $H$, minimum points per slice $n_{\min}$, taper RANSAC residual threshold $\epsilon$, taper slope bound $\kappa$, and~whether solid-circle fitting uses opacity reliability weights ($w_k=\bar{\alpha}_k$) or uniform weights ($w_k\equiv 1$) while keeping all other settings fixed ($\Delta z=0.1$\,m, $K=2000$, $\rho_{\min}=0.1$, $r\in[0.02,1.0]$\,m, and~the tuned radius exponent $p$ from Figure~\ref{fig:power_search}).}

{Table~\ref{tab:ablation} shows that DBH accuracy is stable over a broad range of values. Thin slabs increase error due to sparse support, whereas overly thick slabs can mix residual clutter; $H=1.0$\,m performs best in our data and nearby choices remain competitive. Results are largely insensitive once $n_{\min}\ge 5$ and $\epsilon\ge 2$\,cm. Imposing a moderate taper-slope bound improves robustness, and~disabling this bound increases both RMSE and positive bias. Finally, using opacity as a continuous reliability weight yields a large improvement compared to uniform weighting, confirming the importance of the opacity signal.}

\begin{table}[t]
\centering
\scriptsize
\setlength{\tabcolsep}{2.4pt}
\renewcommand{\arraystretch}{0.94}
\caption{\textbf{Ablation on TreeDGS on the held-out test split (N=189).} Default values are shown in \textbf{bold}. SR is 189/189 for all rows, so only MAE, RMSE, and ME (cm) are shown.}
\label{tab:ablation}
\begin{tabular}{@{}l l r r r@{}}
\toprule
\textbf{Param.} & \textbf{Val.} & \textbf{MAE} & \textbf{RMSE} & \textbf{ME} \\
\midrule
\multirow{5}{*}{\makecell[l]{Slice $H$\\(m)}}
    & 0.2 & 5.19 & 6.62 & $-$3.85 \\
    & 0.5 & 3.86 & 5.23 & $-$1.57 \\
    & \textbf{1.0} & 3.70 & 4.79 & $-$0.38 \\
    & 1.5 & 3.89 & 6.42 & 0.41 \\
    & 2.0 & 4.03 & 5.94 & 0.88 \\
\midrule
\multirow{4}{*}{$n_{\min}$}
    & 3  & 3.94 & 6.35 & $-$0.06 \\
    & \textbf{5}  & 3.70 & 4.79 & $-$0.38 \\
    & 10 & 3.66 & 4.82 & $-$0.36 \\
    & 20 & 3.65 & 4.84 & $-$0.31 \\
\midrule
\multirow{4}{*}{\makecell[l]{$\epsilon$\\(cm)}}
    & 1.0 & 4.15 & 6.63 & $-$0.25 \\
    & \textbf{2.0} & 3.70 & 4.79 & $-$0.38 \\
    & 3.0 & 3.53 & 4.77 & $-$0.41 \\
    & 5.0 & 3.53 & 4.73 & $-$0.27 \\
\midrule
\multirow{5}{*}{\makecell[l]{$\kappa$\\(cm/m)}}
    & 0.1 & 3.78 & 4.94 & $-$0.62 \\
    & 0.2 & 3.71 & 4.87 & $-$0.43 \\
    & \textbf{0.3} & 3.70 & 4.79 & $-$0.38 \\
    & 0.5 & 3.68 & 4.86 & $-$0.10 \\
    & unb. & 5.01 & 7.93 & 2.26 \\
\midrule
\multirow{2}{*}{$w_k$}
    & opacity & 3.70 & 4.79 & $-$0.38 \\
    & uniform & 6.97 & 9.61 & 5.79 \\
\bottomrule
\end{tabular}
\end{table}

\subsubsection{ForestFormer3D Segmentation~Quality}

\begin{figure}
    \centering  
    \includegraphics[width=0.98\linewidth]{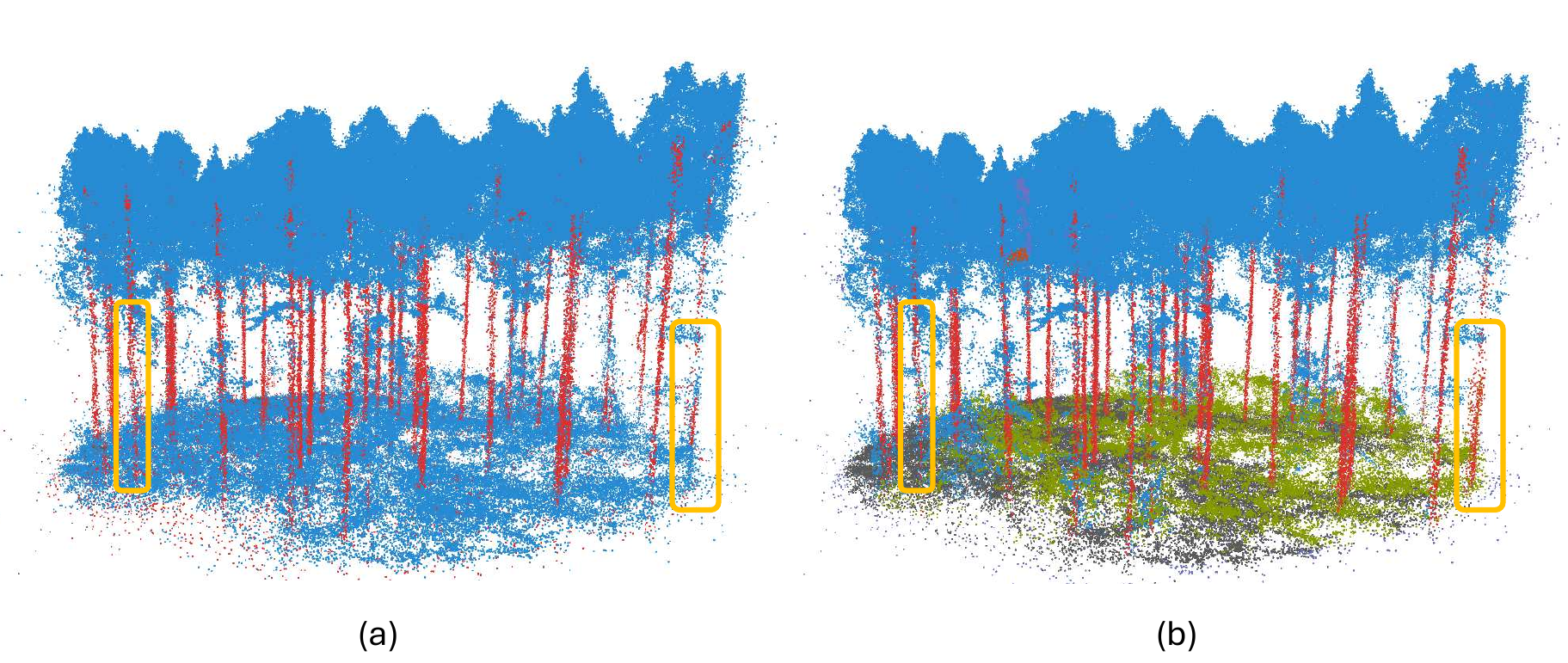}
    \caption{{\textbf{{ForestFormer3D segmentation on Plot~2 and our hand-corrected~reference.}} (\textbf{a}) Off-the-shelf ForestFormer3D trunk/ground predictions and instance IDs (trunk points in red). (\textbf{b})~Hand-corrected labels used for evaluation, obtained by editing the predictions to fix trunk false negatives/positives, ground mislabels, and~instance-ID assignment errors. Creating this reference annotation for Plot~2 required \(\approx\) 14 hours by a {single annotator.} Yellow boxes highlight representative regions where manual edits corrected misclassifications/instance IDs; trunk points are red, while non-trunk points are blue/gray and ground points are green. }}
    \label{fig:forestformer_eval_with_handlablel}
\end{figure}

{Because our field dataset does not provide the point-wise trunk/instance ground truth, we quantify segmentation performance on a small but representative subset (Plot~2) by creating a hand-corrected reference annotation. Starting from the ForestFormer3D predictions, a~single annotator manually fixed trunk false negatives/positives, ground mislabels, and~instance-ID assignment errors for all matched trees in Plot~2 (\(\approx\)14 labeling hours; Figure~\ref{fig:forestformer_eval_with_handlablel}). Table~\ref{tab:forestformer_plot2_metrics} summarizes trunk semantic accuracy (F1 \(=0.812\), IoU \(=0.683\)).}

{
To quantify the impact on DBH, we reran the fitting pipeline on Plot~2 using the hand-corrected labels as an oracle segmentation. This reduces TreeDGS (Cir. w. fit.) RMSE from 5.07 cm to 4.21 cm (MAE from 3.18 cm to 2.68 cm) on the same 22 matched trees (Table~\ref{tab:plot2_oracle_seg}). This indicates that segmentation errors contribute to DBH error, but~a substantial portion of the remaining error is due to reconstruction/visibility limitations rather than segmentation alone.
}

\begin{table}
\centering
\scriptsize
\setlength{\tabcolsep}{4pt}
\renewcommand{\arraystretch}{0.98}
\caption{\textbf{ForestFormer3D segmentation accuracy on Plot~2.} Hand-corrected reference; trunk semantic segmentation over 2,396,986 points.}
\label{tab:forestformer_plot2_metrics}
\begin{tabular}{l c c}
\toprule
\textbf{Metric} & \textbf{Value} & \textbf{Notes} \\
\midrule
Precision & 0.760 & TP=107,746; FP=34,038 \\
Recall    & 0.871 & FN=15,974 \\
F1        & 0.812 & --- \\
IoU       & 0.683 & --- \\
Accuracy  & 0.979 & TN=2,239,228 \\
\bottomrule
\end{tabular}
\end{table}

\begin{table}[t]
\centering
\scriptsize
\setlength{\tabcolsep}{2.6pt}
\renewcommand{\arraystretch}{0.98}
\caption{\textbf{Effect of segmentation quality on DBH accuracy (Plot~2, 22 trees).} TreeDGS (Cir.\ w.\ fit.) with ForestFormer3D predictions versus hand-corrected oracle labels. MAE/RMSE/ME are in cm and RRMSE is in \%.}
\label{tab:plot2_oracle_seg}
\begin{tabular}{@{}l c c c c c@{}}
\toprule
\textbf{Segmentation} & \textbf{SR} & \textbf{MAE} & \textbf{RMSE} & \textbf{RRMSE} & \textbf{ME} \\
\midrule
ForestFormer3D & 22/22 & 3.18 & 5.07 & 17.82 & 1.88 \\
Oracle         & 22/22 & 2.68 & 4.21 & 14.81 & 1.49 \\
\bottomrule
\end{tabular}
\end{table}

\subsubsection{Runtime~Analysis}
\label{subsec:runtime}

{To assess practical utility, we report a representative runtime breakdown for one plot (Plot~2, 365 RGB images) processed end to end on a single workstation (AMD Ryzen Threadripper PRO 7995WX; 192 CPU threads; 1$\times$ NVIDIA RTX 5090, 32\,GB). Reported times are wall-clock and depend on image count, scene difficulty, and~hardware, but~they indicate that the overall cost is dominated by SfM matching and Gaussian optimization, while DBH fitting itself is negligible (Table \ref{tab:runtime_breakdown})}

\begin{table}[t]
\centering
\scriptsize
\setlength{\tabcolsep}{3pt}
\renewcommand{\arraystretch}{0.98}
\caption{\textbf{Runtime breakdown per plot (Plot~2, 365 images).} Feature matching and Gaussian optimization dominate runtime; segmentation and DBH fitting are comparatively fast.}
\label{tab:runtime_breakdown}
\begin{tabularx}{\columnwidth}{@{}X c c@{}}
\toprule
\textbf{Stage} & \textbf{Time} & \textbf{Resource} \\
\midrule
Feature matching (TopicFM) & 50 min & GPU \\
SfM + bundle adjustment (GLOMAP) & 3 min & CPU \\
OpenMVS densification & 5 min & CPU \\
Gaussian optimization (RaDe-GS) & 20 min & GPU \\
ForestFormer3D segmentation & 5 min & GPU \\
Slice fitting + taper aggregation & 10 s & CPU \\
\midrule
Total & \(\approx\)83 min & --- \\
\bottomrule
\end{tabularx}
\end{table}

{Since plots are independent, the~pipeline is trivially parallelizable across plots (one plot per GPU).}

\section{Discussion}

Overall, our results suggest that TreeDGS can convert pixel-limited UAV RGB imagery into a trunk-centric 3D representation that is sufficiently coherent for plot-scale DBH estimation, narrowing the performance gap to a UAV LiDAR baseline while using only low-cost imagery. Compared to conventional SfM/MVS densification, Gaussian optimization combined with opacity-guided sampling produces denser and more spatially consistent support near breast height, which directly benefits downstream cross-sectional fitting. The~added breast-height observability analysis further indicates that DBH errors increase primarily in the low-visibility regime, reinforcing that view scarcity and occlusion are key bottlenecks in aerial DBH measurement. In~addition, our Plot~2 hand-correction study highlights that segmentation quality contributes measurably to DBH accuracy, but~this does not fully account for residual error, implying that reconstruction completeness and boundary visibility remain limiting factors even with improved labels. Finally, the~runtime breakdown shows that the full pipeline is practically feasible at the plot level, and~that the proposed slice fitting stage is computationally negligible compared to reconstruction and learning-based components. These findings collectively support TreeDGS as a practical RGB-only alternative (or complement) for rapid forest inventory under appropriate visibility conditions. However, several limitations remain.

\begin{figure*}
    \centering
    \includegraphics[width=0.98\linewidth]{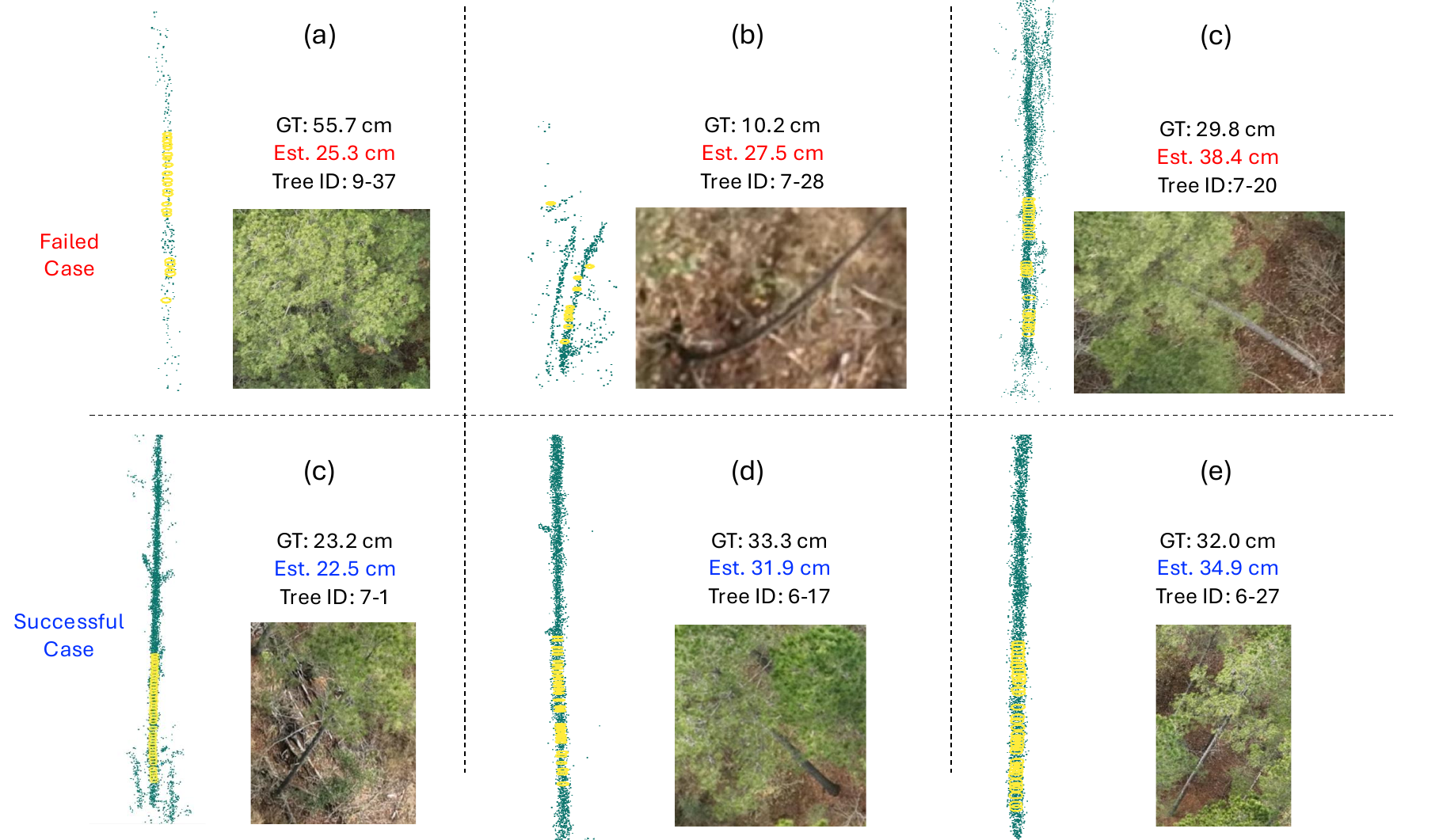}
    \caption{{\textbf{{Examples of successful and failed DBH fits at breast~height.}} (\textbf{a}--\textbf{c}) Failure cases: (\textbf{a}) no unoccluded breast-height views, (\textbf{b}) strongly curved/leaning stem causing biased horizontal slicing, and~(\textbf{c}) instance segmentation that merges a target stem with a nearby sapling. (\textbf{d}--\textbf{f}) Successful cases with clear breast-height visibility and accurate per-slice fits. Yellow circles show the fitted cross-sections on {representative slices.} Green dots show the segmented trunk points; the estimated DBH ({Est.}) is colored red for failure cases and blue for successful cases. }}
    \label{fig:Success_and_Failed_Trees}
\end{figure*}

\paragraph{Limitations.} TreeDGS has important limitations.
First, it relies on trunk visibility: the breast-height band must be observed in at least a few frames (in practice, \(\sim\)2+ informative views
for SfM/MVS and Gaussian optimization to constrain the stem surface {(Figure \ref{fig:Success_and_Failed_Trees}a)}.
When this band is fully occluded in most views, the~method cannot recover the missing geometry, regardless of how densely the Gaussian field is sampled; sampling density cannot compensate for geometry that is never sufficiently observed in the input~imagery.

Second, DBH estimation is sensitive to 3D semantic segmentation quality.
Because diameter is fit from trunk-isolated points, false-positive trunk labels from nearby vegetation can bias the estimate {(Figure \ref{fig:Success_and_Failed_Trees}c)}.
Although multi-view opacity weighting mitigates this effect, the~pipeline remains subject to the failure modes of the trunk classifier~\cite{xiang2025forestformer3d}.
This sensitivity can also contribute to poorer LiDAR baseline performance in challenging plots: when breast-height stem returns a small number of vegetation false positives can disproportionately bias cylinder~fits. 

{Third, our current DBH estimator makes a morphological assumption that is well satisfied in the managed Pinus taeda plantation used here but may not hold in more complex forests. Specifically, we assume that (i) the lower stem cross-section can be reasonably approximated by a circle (per-slice solid-circle RANSAC), (ii) horizontal slicing is a good proxy for slicing orthogonal to the stem axis (i.e., stems are approximately vertical/straight near breast height), and~(iii) diameter decreases monotonically with height over the fitted lower-stem segment (linear taper with a negative-slope prior). In~natural or mixed-species forests, stems can be leaning, curved (Figure \ref{fig:Success_and_Failed_Trees}b), buttressed, fluted, forked, or~multi-stem, and~these cases can violate the above assumptions and introduce bias. Importantly, this limitation is in the geometric measurement model rather than in TreeDGS sampling itself: the opacity-weighted sampling yields a denser, reliability-weighted point set that could support more flexible trunk models. We view extending the measurement stage to such morphologies and validating on mixed-species natural forests as key future work.}

\paragraph{Future Work.} A promising direction is to reduce dependence on post-processing (sampling and segmentation) by incorporating explicit trunk primitives into Gaussian optimization.
Jointly optimizing a small set of parameterized stem elements (e.g., tapered cylinders) together with the Gaussian field would enable DBH to be inferred directly from model parameters and would provide a stronger inductive bias under partial trunk visibility.
Such a hybrid representation may also support end-to-end supervision when ground-truth field DBH is~available.

\section{Conclusions}

We introduced TreeDGS, a~UAV RGB-only pipeline that repurposes 3D Gaussian splatting as an opacity-weighted density sampling method for trunk measurement, rather than only a view-synthesis renderer. Building on SfM/MVS initialization, TreeDGS optimizes a Gaussian field and then converts it into a measurement-ready stem point set via depth-aware cumulative-opacity sampling and multi-view reliability weighting, enabling robust circle fitting at breast height. In~doing so, our approach bridges a key gap left by prior UAV LiDAR and UAV-SfM DBH pipelines that operate on inherently sparse or view-limited point clouds, and~demonstrates that Gaussian splats can support accurate trunk-level metrics from commodity aerial~imagery.

\vspace{+6pt}

\section*{Acknowledgments}
We thank PotlatchDeltic Corporation for supporting this research by allowing us to conduct data collection and validation on their managed forest stands. We are grateful to Jacob Strunk, Kit Hart, Nathaniel Naumann, Paurava Thakore, Tim Sydor, and Bill Driscoll for coordinating logistics, sharing field context, and assisting with site access and study planning.

\section*{Data Availability}
{The original contributions presented in this study are included in the article. Further inquiries can be directed to the corresponding author.}

\section*{Conflict of Interest}
Authors Belal Shaheen, Minh-Hieu Nguyen, Bach-Thuan Bui, Shubham, Tim Wu, Michael Fairley, Matthew Zane, and Michael Wu were employed by Coolant. \textcolor{blue}{James Tompkin is a scientific advisor to Coolant.} \textcolor{blue}{The authors declare that these affiliations did not influence the study design, data collection, analysis, interpretation of results, or the decision to publish.}

{\small
\bibliographystyle{ieeenat_fullname}
\bibliography{main}
}
\end{document}